\documentclass[runningheads]{llncs}
\usepackage{makeidx}
\usepackage{graphicx}
\usepackage{amsmath,amssymb} % define this before the line numbering.
\usepackage{color}
\usepackage{subcaption}
\usepackage{etoolbox}
\makeatletter
\preto{\@verbatim}{\topsep=0pt \partopsep=0pt }
\makeatother

\begin{document}
	\pagestyle{headings}
	\mainmatter

	% Insert your submission number here
	\def\GCPR20SubNumber{104}

	% Replace with your title
	\title{Multimodal semantic forecasting based on conditional generation of 
	future features\thanks{This work has been funded
	  by Rimac Automobili and supported 
	  by the European Regional Development Fund 
	  under the grant KK.01.1.1.01.0009 DATACROSS.}}

	% DO NOT MODIFY these for the draft version that is used for the
	% review process.
	\titlerunning{Multimodal semantic forecasting based on conditional generation of future...}
	\authorrunning{Kristijan Fugošić, Josip Šarić and Siniša Šegvić}
	\author{Kristijan Fugošić, Josip Šarić and Siniša Šegvić}
	\institute{University of Zagreb Faculty of Electrical Engineering and Computing, Croatia}

	\maketitle

	\begin{abstract}
        This paper considers semantic forecasting in road-driving scenes.
        Most existing approaches address this problem
        as deterministic regression of future features
        or future predictions given observed frames.
        However, such approaches ignore the fact
        that future can not always be guessed with certainty.
        For example, when a car is about 
        to turn around a corner,
        the road which is currently occluded by buildings
        may turn out to be either free to drive, 
        or occupied by people, other vehicles or roadworks.
        When a deterministic model confronts such situation,
        its best guess is to forecast the most likely outcome.
        However, this is not acceptable since it 
        defeats the purpose of forecasting 
        to improve security.
        It also throws away valuable training data,
        since a deterministic model is unable 
        to learn any deviation from the norm.
        We address this problem by providing 
        more freedom to the model through allowing it
        to forecast different futures.
        We propose to formulate multimodal forecasting 
        as sampling of a multimodal generative model 
        conditioned on the observed frames.
        Experiments on the Cityscapes dataset
        reveal that our multimodal model 
        outperforms its deterministic counterpart 
        in short-term forecasting while 
        performing slightly worse in the mid-term case.
	\end{abstract}

    \section{Introduction}
    Self-driving cars are today's 
    burning topic \cite{yao19icra}. 
    With their arrival, the way that we look at passenger and freight traffic will change forever. 
    But in order to solve such a complex task, we must first solve a series of "simpler" problems. 
    One of the most important elements of an autonomous driving system is the ability to recognize and understand the environment
    \cite{kirillov2018panoptic,chen20arxiv}. 
    It is very important that the system is able 
    to recognize roads, pedestrians moving along or on the pavement, other cars and all other traffic participants.
    This makes semantic segmentation 
    a very popular problem
    \cite{zhao17cvpr,yang18cvpr,zhen19aaai}. 
    
    However, the ability to predict the future 
    is an even more important attribute 
    of intelligent behavior
    \cite{vondrick2015anticipating,vukotic2017one,luc2017predicting,su2017predicting,terwilliger2019recurrent}. 
    It is intuitively clear that critical real-time systems
    such as autonomous driving controllers
    could immensely benefit from the ability 
    to predict the future by considering the past
    \cite{yao19icra,rprior,probabilistic}.
    Such systems could make much better decisions 
    than their counterparts which are able 
    to perceive only the current moment.
    Unfortunately, this turns out to be a very hard problem.
    Most of the current work in the field
    approaches it very conservatively, 
    by forecasting only unimodal future
    \cite{luc2017predicting,sun19mm}.
    However, this approach makes 
    an unrealistic assumption
    that the future is 
    completely determined by the past,
    which makes it suitable
    for guessing only the short-term future.
    Hence, deterministic forecasting approaches
    will be prone to allocate most of its forecasts  
    to instances of common large classes 
    such as cars, roads, sky and similar.
    On the other side, such approaches will often 
    underrepresent smaller objects. When it comes to 
    signs, poles, pedestrians or some other thin objects, 
    it makes more sense for a conservative model to 
    allocate more space to the background than to 
    risk classifying them. 
    Additionally, future locations 
    of dynamic and articulated objects
    such as pedestrians or domestic animals 
    would also be very hard to forecast
    by a deterministic approach. 
    
    In order to address problems of unimodal forecasting, 
    this work explores how to 
    equip a given forecasting model 
    with somewhat more freedom, 
    by allowing and encouraging 
    prediction of different futures. 
    Another motivation for doing so involves scenarios 
    where previously unseen space is unoccluded. 
    Such scenarios can happen when we are 
    turning around a corner or when another car 
    or some larger vehicle is passing by. 
    Sometimes we can deduce what could be 
    in that new space by observing recent past, 
    and sometimes we simply can't know. 
    In both cases, we would like our model 
    to produce a distribution 
    over all possible outcomes
    in a stochastic environment 
    \cite{bhattacharyya2018bayesian,kosaraju19nips,rprior,probabilistic}.
    We will address this goal 
    by converting the basic regression model 
    into a conditional generative model
    based on adversarial learning 
    \cite{cgan}
    and moment reconstruction losses 
    \cite{mrgan}.

    \section{Related work}
    \subsubsection{Dense semantic forecasting.} 
    Predicting future scene semantics 
    is a prominent way 
    to improve accuracy and reaction speed 
    of autonomous driving systems.
    Recent work shows that 
    direct semantic forecasting 
    is more effective than 
    RGB forecasting \cite{luc2017predicting}.
    Further work proposes to forecast 
    features from an FPN pyramid
    by multiple feature-to-feature (F2F) 
    models \cite{f2f}.
    This has recently been improved
    by single-level F2F forecasting 
    with deformable convolutions 
    \cite{saric,saric20cvpr}.
    
    \subsubsection{Multimodal forecasting.} 
    Future is uncertain and multimodal, 
    especially in 
    %mid-term and 
    long-term forecasting. 
    Hence, foreasting multiple futures 
    is an interesting research goal. 
    An interesting related work
    forecasts multi-modal 
    pedestrian trajectories
    \cite{kosaraju19nips}.
    Similar to our work,
    they also achieve
    multimodality through a
    conditional GAN framework.
    Multi-modality has also been expressed
    through mixture density networks \cite{rprior} 
    in order to forecast 
    egocentric localization
    and emergence prediction.
    None of these two works 
    consider semantic forecasting.

    To the best of our knowledge, 
    there are only a few works
    in multimodal semantic forecasting,
    and all these works 
    are either very recent \cite{bhattacharyya2018bayesian}
    or concurrent \cite{rprior}. 
    One way to address multimodal 
    semantic forecasting
    is to express inference 
    within a Bayesian framework \cite{bhattacharyya2018bayesian}.
    However, Bayesian methods are known 
    for slow inference and 
    poor real-time performance.
    Multi-modality can also be expressed
    within a conditional 
    variational framework
    \cite{probabilistic}, 
    by modelling interaction 
    between the static scene, 
    moving objects and multiple 
    moving objects.
    However, the reported performance
    suggests that the task is far 
    from being solved.

    \subsubsection{GANs with moment reconstruction.} 
    GANs \cite{gan} and their 
    conditional versions \cite{cgan}
    have been used in many tasks
    \cite{pix2pix,inpaintning,supres,vidpred}. 
    However, these approaches lack output diversity 
    due to mode collapse. 
    Recent work alleviates this problem 
    with moment reconstruction loss \cite{mrgan} 
    which also improves the training stability. 
    
    \subsubsection{Improving semantic segmentation with adversarial loss.} 
    While most GAN discriminators
    operate on raw image level, 
    they can also be applied 
    to probabilistic maps. 
    This can be used either as 
    a standalone loss \cite{prostate}
    or as a regularizer
    of the standard cross entropy loss
    \cite{adlearning}.
    
    \section{Method}
        
    \subsection{Conditional MR-GAN}
        Generative adversarial models \cite{gan} are comprised of two neural networks - a generator and a discriminator. Each of them has its own task and separate loss function. The goal of a generator is to produce diverse and realistic samples, while discriminator classifies given sample as either real (drawn from the dataset) or fake (generated).
        By conditioning the model it is possible to direct the data generation process. 
        Generative adversarial networks can be extended to a conditional model if both the generator and discriminator are conditioned on some additional information \cite{cgan}. Additional information can be of any kind, in our case it's a blend of features extracted from past frames.
        
        However, both standard GAN and its conditional version are highly unstable to train. 
        To counter the instability, 
        most conditional GANs
        for image-to-image translation \cite{pix2pix}
        use reconstruction (l1/l2) loss 
        in addition to the GAN loss. 
        While reconstruction loss forces model 
        to generate samples similar to ground-truth,
        it often results in \textit{mode collapse}.
        Mode collapse is one of the greatest problems
        of generative adversarial models.  
        While we desire diverse outputs, 
        mode collapse manifests itself 
        as one-to-one mapping.
        This problem can be mitigated by replacing 
        the traditional reconstruction loss with
        \textit{moment reconstruction (MR)} losses
        which increase training stability 
        and favour multimodal output generation
        \cite{mrgan}.
        
        The main idea of MR-GAN \cite{mrgan} is 
        to use maximum likelihood estimation loss 
        to predict conditional statistics 
        of the real data distribution. 
        Specifically, MR-GAN estimates 
        the central measure and the dispersion
        of the underlying distribution,
        which correspond to mean and variance 
        in the Gaussian case.
        % (TODO or which are mean and var...)
        
        %\vspace{2mm}
    
        \begin{equation}
        \mathcal{L}_{MLE,Gaussian}=\mathbb{E}_{x,y}\left[ \frac{(y- \hat{\mu})^2}{2\hat{\sigma}^2} + \frac{1}{2}\log\hat{\sigma}^2 \right], \textrm{ where }(\hat{\mu},\hat{\sigma}^2) = f_\theta (x)
        \label{eq:gaussian}
        \end{equation}
        
        %\vspace{2mm}
        
        Overall architecture of MR-GAN is similar to conditional GANs, with two important novelties:
        \begin{enumerate}
        \item 
          Generator produces $K$ different samples $\hat{y}_{1:K}$ for each image $x$ by varying random noise $z_{1:K}$. \\
        \item 
          Loss function is applied 
          to the sampled moments (mean and variance)
          in contrast to the reconstruction loss 
          which is applied directly on the samples.\\
        \end{enumerate}
        They estimate the moments of the generated distribution as follows:
        
        %\vspace{2mm}
        \begin{equation}
        \tilde{\mu}=\frac{1}{K}\sum_{i=1}^{K}\tilde{y}_i, \textrm{  }
        \tilde{\sigma}^2=\frac{1}{K-1}\sum_{i=1}^{K}(\tilde{y}_i-\tilde{\mu})^2, \textrm{ where }
        \tilde{y}_{1:K}=G(x,z_{1:K}).
        \end{equation}
        %\vspace{2mm}
        
        MR loss is calculated by plugging $\tilde{\mu}$ and $\tilde{\sigma}^2$ in Eq. \ref{eq:gaussian}. The loss thus obtained is called MR2, while they denote a loss that does not take into account variance with MR1.
        
        For more stable learning, especially at an early stage, the authors suggest a loss called \textit{Proxy Moment Reconstruction (proxy MR)} loss. 
        As It was shown in \cite{mrgan} that MR and proxy MR losses achieve similar results on Pix2Pix \cite{pix2pix} problem, 
        we will use simpler MR losses for easier, \textit{end-to-end}, training.

    \subsection{F2F forecasting}
        Most of the previous work in forecasting focuses on predicting raw RGB future frames and subsequent semantic segmentation. Success in that area would be a significant achievement because it would make possible to train on extremely large set of unmarked learning data. However, problems such as autonomous driving require the program to recognize the environment on a semantically meaningful level. In that sense forecasting on RGB level is an unnecessary complication. 
        As many attempts in feature-to-feature forecasting were based on semantic segmentation, in \cite{f2f} they go a step further and predict the semantic future at the instance level. This step facilitates understanding and prediction of individual objects trajectories. The proposed model shares much of the architecture with the Mask R-CNN, with the addition of predicting future frames. Since the number of objects in the images varies, they do not predict the labels of the objects directly. Instead, they predict convolutional features of fixed dimensions. Those features are then passed through the detection head and upsampling path to get final predictions.
        %TODO bolje slozit recenice

    %\newpage        
    \subsection{Single-level F2F forecasting}
        Šarić et al. in their paper \cite{saric} proposed a single-level F2F model with deformable convolutions. The proposed model, denoted as DeformF2F, brings few notable changes compared to \cite{f2f}:
        \begin{enumerate}
        \item Single-level F2F model which performs on last, spatially smallest, resolution
        \item Deformable convolutions instead of classic or dilated ones
        \item Ability to fine tune two separately trained submodels (F2F and submodel for semantic segmentation).
        \end{enumerate}
        
        DeformF2F achieves \textit{state-of-the-art} performance on mid-term (t+9) prediction, and second best result on short-term (t+3) prediction.

    \subsection{Multimodal F2F Forecasting}
    
    	   We use modified single-level F2F forecasting model as a generator, customized PatchGAN \cite{patchgan} as a discriminator, while MR1 and MR2 losses are used in order to achieve diversity in predictions. We denote our model as MM-DeformF2F (Multimodal DeformF2F).
    	
    	\begin{figure}[htb]
        \centering
        \includegraphics[width=.85\textwidth]{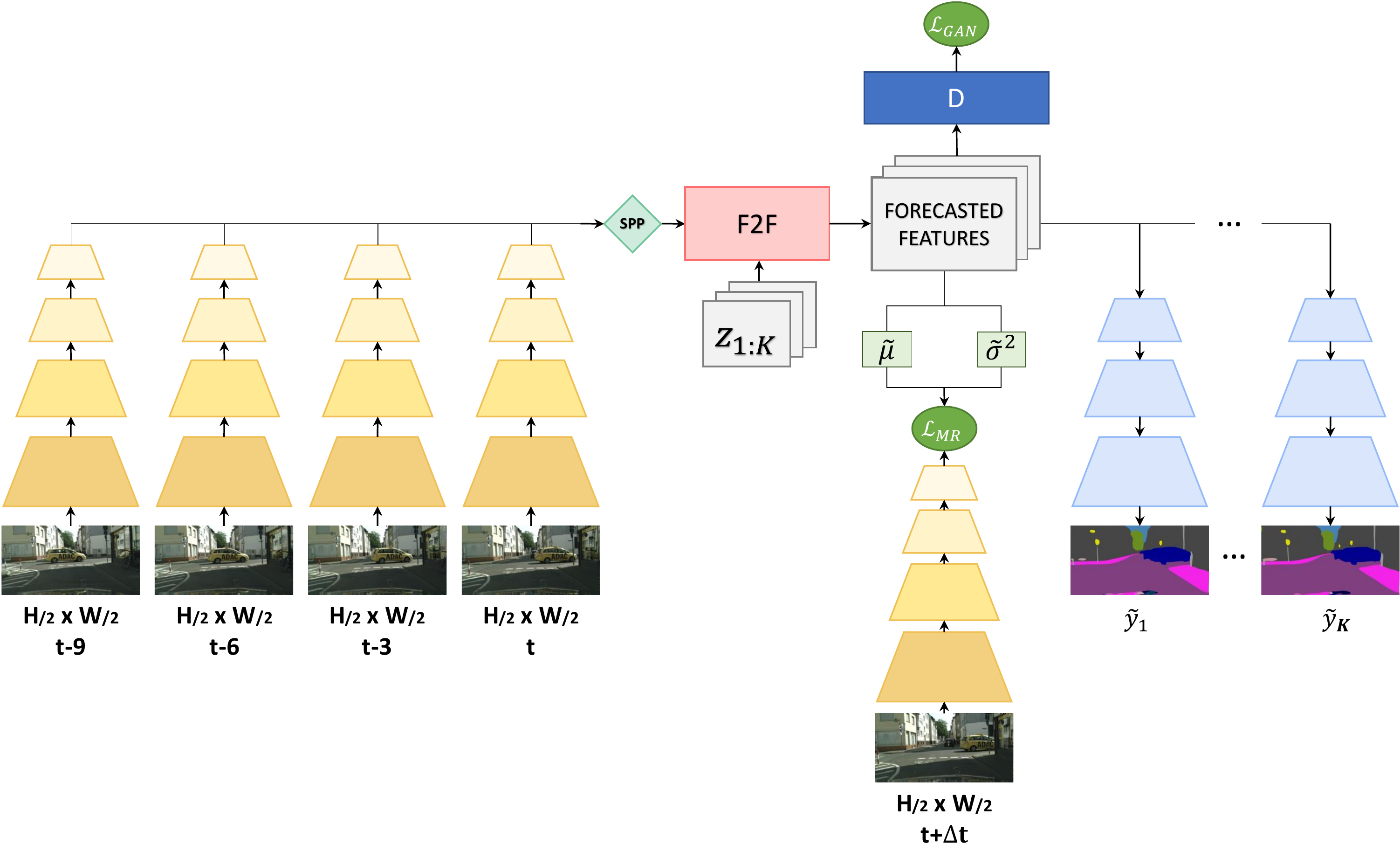}
        \caption{Structural diagram of our model. Base structure is similar to \cite{saric}. It is composed of feature extractor (yellow), upsampling branch (blue), spatial pyramid pooling (SPP) and F2F. Additionally, we introduce random noise $z$, discriminator $D$ and new loss functions.  Our model generates multiple predictions. }
        \label{fig:mmf2f}
        \end{figure}
    	
    	\subsubsection{Generator.}
    	    Generator is based on DeformF2F model. In order to generate diverse predictions, we introduce noise in each forward pass. Gaussian noise tensor has 32 channels and fits the spatial dimensions of the input tensor. Instead of one, we now generate $K$ different predictions with the use of $K$ different noise tensors. Generator is  trained with MR and GAN loss applied to those predictions. 
    	    
    	\subsubsection{Discriminator.}
    		According to the proposal from \cite{mrgan}, we use PatchGAN as a discriminator. Since input features of our PatchGAN are of significantly smaller spatial dimensions, we use it in a modified form with a smaller number of convolutional layers. Its purpose is still to reduce the features to smaller regions, and then to judge each region as either fake (generated) or real (from dataset). Decisions across all patches are averaged in order to bring final judgment in the form of 0 to 1 scalar. Since the discriminator was too dominant in learning, we introduced dropout in its first convolutional layer. In general, we shut down between 50 and 65 percent of features.

    	\subsubsection{Dataset.}    	
    		Following the example of \cite{saric}, we use video sequences from the Cityscapes dataset. The set contains 2975 scenes (video sequences) for learning, 500 for validation and 1525 for testing with labels for 19 classes. Each scene is described with 30 images, with a total duration of 1.8 seconds. That means that dataset contains a total of 150,000 images with resolution $1024 \times 2048$ pixels. Ground-truth semantic segmentation is available for the 20th image of each scene. Since introduction of GAN methods made our model more complicated, all images from the dataset were halved in width and height in order to reduce the number of features and speed up training. 
    		
    	\subsubsection{Training procedure.}
			In last paragraph we described the dataset and the initial processing of the input data. If we denote the current moment as $t$, then in short-term prediction we use convolutional features at moments $t-9$, $t-6$, $t-3$ in order to predict the semantic segmentation at moment $t+3$, or at moment $t+9$ for mid-term forecasting. 
			Features have spatial dimensions $16 \times 32$ and 128 channels. Training can be divided into two parts. First, we jointly train feature extractor and upsampling branch with cross entropy loss \cite{saric,ladder}. All images later used for training are passed through feature extracting branch and the resulting features are stored on SSD drive. We later load those features instead of passing through feature extractor, as that saves us time in successive training and evaluation of the model.
            In the second part, we train the F2F model in an unsupervised manner. Unlike \cite{saric}, instead of L2 loss we use MR loss and GAN loss. We give a slightly greater influence to the reconstruction loss ($\lambda_{MR}=100$) than adversarial ($\lambda_{GAN}=10$). For both the generator and the discriminator, we use Adam optimizer with a learning rate of $4 \cdot 10^{-4}$ and decay rates 0.9 and 0.99 for the first and second moment estimates, respectively. We reduce the learning rate using cosine annealing without restart to a minimum value of $1 \cdot 10^{-7}$. To balance  the generator and the discriminator, we introduce dropout in first convolutional layer of the discriminator. As an example, training short-term forecasting task with MR1 loss without dropout begins to stagnate as early as the fortieth epoch, where mIoU is 1.5 to 2 percentage points less than the best results achieved.

    \section{Experiments}
    	We show average metrics across 3 trained models and multiple evaluations for each task. In every forward pass we generate 8 predictions. We use mIoU as our main metric for accuracy, while MSE and LPIPS are used to express diversity as explained below.

       	    \subsubsection{MSE} 
       	    Mean Squared Error is our main diversity metric.
       	    We measure Euclidean distance on pixel level between every two generated predictions for each scene, and take mean over whole dataset.
			\subsubsection{LPIPS} 
            Following the example of \cite{mrgan}, we also use LPIPS (\textit{Learned Perceptual Image Patch Similarity} \cite{lpips})  to quantify the diversity of generated images. In \cite{lpips} they have shown that deep features can be used to describe similarity between two images while outperforming  traditional measures like L2 or SSIM. We measure LPIPS between every two generated predictions for each scene, and take mean over whole dataset. Since we don't generate RGB images, but instead predict semantic future which has limited structure, MSE has proved to be sufficient measure for diversity.

    	\subsection{Visual assessment} 
    		In addition to the numerical results, in following subsections we will also show generated predictions, as well as two gray images:\\
            \indent a) Mean logit variance 
            \begin{verbatim}
            logits.var(dim=0).mean(dim=0)
            \end{verbatim}
            
            b) Variance of discrete predictions
            
            \begin{verbatim}
            logits.argmax(dim=1).double().var(dim=0)
            \end{verbatim}
            
\begin{figure}[htb]
            \includegraphics[width=1\linewidth]{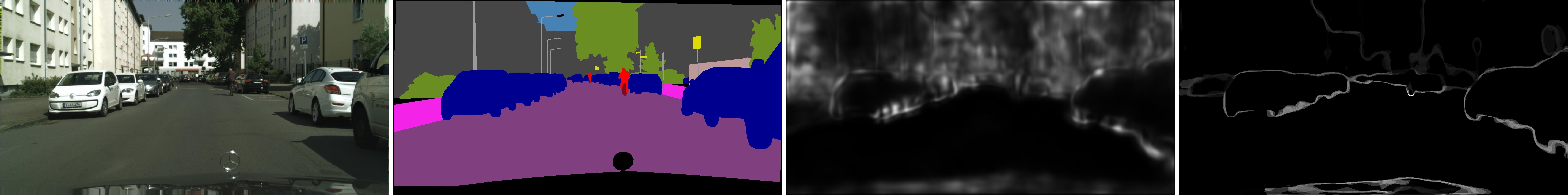}
            \caption{Shown in the following order: future frame and its ground truth segmentation, mean logit variance and variance of discrete predictions. The first gray image highlights areas of uncertainty, while on the second gray image we see areas that are classified into different classes on different generated samples. The  higher the uncertainty, the whiter the area.}
            \label{fig:gray}
            \end{figure}

            An example of gray images is shown in Figure \ref{fig:gray}. The first gray image highlights areas of uncertainty, while on the second image we observe areas that are classified into different classes on different generated samples.
            
    	\subsection{Experimental results on Cityscapes dataset}
\begin{figure}[htb]
        	\centering
            \includegraphics[width=0.95\linewidth]{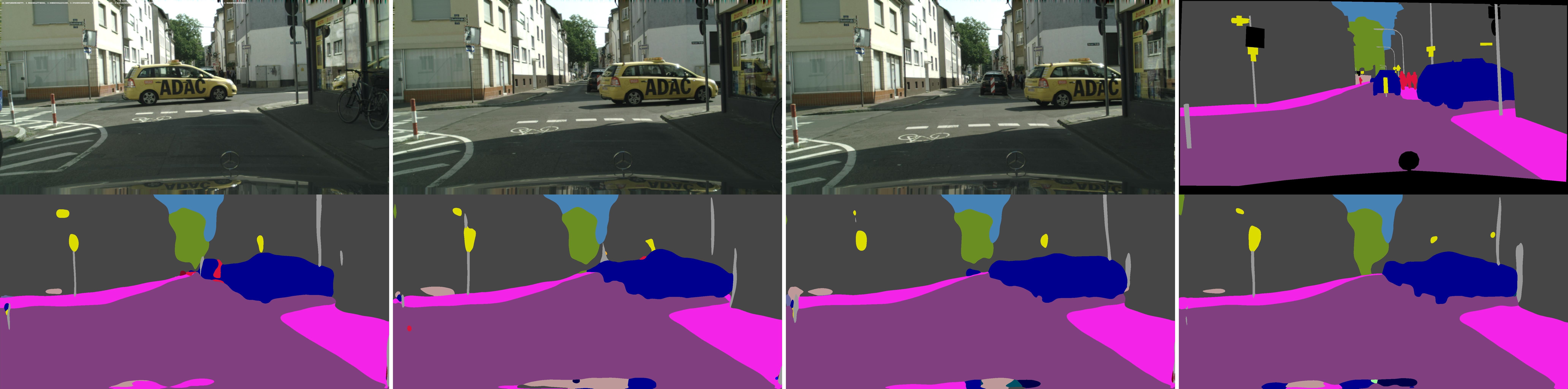}
            \caption{Short-term forecasting with MR2 loss. 
              Row 1 shows the first and the last input frame, 
              the future frame, and its ground-truth segmentation. 
              Row 2 shows 4 out of 12 model predictions. 
            }
            \label{fig:res}
            \end{figure}

        	We conducted experiments on short-term and mid-term forecasting tasks with roughly the same hyperparameters. 
            With  MR1  loss  we  observe  mIoU  which  is  on  par  with  baseline  model, and slightly greater in the case of short-term forecasting. On the other hand, using MR2 loss resulted in lower mIoU, but predictions are a lot more diverse compared to MR1. 
            Although we could get higher mIoU on mid-term forecasting with minimal changes in hyperparameters at the cost of diversity, we do not intervene because mIoU is not the only relevant measure in this task.  
            Accordingly, although MR2 lowers mIoU by 5 or more percentage points, we still use it because of the greater variety. Visually most interesting predictions were obtained by using MR2 on short-term forecasting task. One of those is shown on Figure \ref{fig:res}. Notice that people are visible in the first frame and obscured by the car in the last frame. In the future moment, the car reveals the space behind it, and for the first time our model predicts people in correct place (second row, first prediction). We failed to achieve something like that when using MR1 loss or with baseline model. Such predictions are possible with the MR2 loss, but still rare, as in this particular case model recognized people in the right place in only one of the twelve predictions.

        	Our main results are shown in tables \ref{tab:resshort} and \ref{tab:resmid}. Since results in \cite{saric} were obtained on images of full resolution, we retrained their model on images with halved height and width. In tables we show average mIoU and mIoU-MO (\textbf{M}oving \textbf{O}bjects) across five different models. We also show results achieved with Oracle,  single-frame model used to train the feature extractor and the upsampling path, which "predicts" future segmentation by observing a future frame. While oracle represents upper limit, \textit{Copy last segmentation} can be seen as lower bound, or as a good difficulty measure for this task. We get a slightly better mIoU if we average the predictions, although this contradicts the original idea of this paper. Like in \cite{bhattacharyya2018bayesian}, we also observe slight increase of mIoU when comparing top 5\% to averaged predictions. Performance boost is best seen when we look at moving objects accuracy (mIoU-MO) while using MR1 loss, as we show in more detail in supplementary. 
        	
            \begin{table}[htb]
                \centering
                \def\arraystretch{1.2}
                \setlength{\tabcolsep}{0.7em} 
                \begin{tabular}{ |c|c|c|c|c|c| } 
                \hline
                Method & Loss & mIoU & mIoU-MO & MSE & LPIPS \\
                \hline
                Oracle & L2 & 66.12 & 64.24 &/ & /\\
                Copy last segmentation & L2 & 49.87 & 45.63 & / & /\\
                \hline
                DeformF2F-8\cite{saric} & L2 & 58.98$^{\pm 0.17}$ & 56.00$^{\pm 0.20}$ & / & /\\
                \hline
                MM-DeformF2F-8 & MR1 & 59.22$^{\pm 0.05}$ & 56.40$^{\pm 0.08}$ & 1.21$^{\pm  0.05}$ & 0.0482 \\
                MM-DeformF2F-8 avg & MR1 & 59.46$^{\pm 0.06}$ & 56.66$^{\pm 0.09}$ & / & / \\
                MM-DeformF2F-8 & MR2 & 53.85$^{\pm 0.35}$ & 49.52$^{\pm 0.64}$ & 4.38$^{\pm  0.24}$ & 0.1519 \\ 
                MM-DeformF2F-8 avg & MR2 & 56.81$^{\pm 0.20}$ & 53.38$^{\pm 0.43}$ & / & / \\
                \hline
                \end{tabular}
                \vspace{1mm}
                \caption{Short-term prediction results. While our model achieves slightly higher mIoU with MR1 loss, MR2 results in much more diverse predictions.}
                \label{tab:resshort}
            \end{table}
            
            %\vspace{5mm}
            
            \begin{table} [htb] 
                \centering
                \def\arraystretch{1.2}
                \setlength{\tabcolsep}{0.7em}           
                \begin{tabular}{ |c|c|c|c|c|c| } 
                \hline
                Method & Loss & mIoU & mIoU-MO & MSE & LPIPS \\
                \hline
                Oracle & L2 & 66.12 & 64.24 & / & /\\
                Copy last segmentation & L2 & 37.24 & 28.31 & / & /\\
                \hline
                DeformF2F-8\cite{saric} & L2 & 46.36$^{\pm 0.44}$ & 40.78$^{\pm 0.99}$ & / & /\\
                \hline
                MM-DeformF2F-8 & MR1  & 46.23$^{\pm 0.28}$ & 41.07$^{\pm 0.57}$ & 2.81$^{\pm 0.32}$ & 0.1049\\
                MM-DeformF2F-8 avg & MR1 & 46.96$^{\pm 0.21}$ & 41.90$^{\pm 0.53}$ & / & /\\
                MM-DeformF2F-8 & MR2 & 37.48$^{\pm 0.31}$ & 29.66$^{\pm 0.60}$ & 7.42$^{\pm 0.44}$ & 0.2279\\
                MM-DeformF2F-8 avg & MR2 & 40.32$^{\pm 0.12}$ & 32.42$^{\pm 0.37}$ & / & /\\
                \hline
                \end{tabular}
                \vspace{1mm}
                \caption{Mid-term prediction results. While our model achieves slightly lower mIoU with MR1 loss, MR2 results in much more diverse predictions.}
                \label{tab:resmid}
            \end{table}
            %\vfill

    	\subsection{Impact of the number of predictions on performance} 
    		Table 10.3 shows the impact of the number of generated predictions ($K$) on their diversity and measured mIoU. We can see that larger number of generated predictions contributes to greater diversity, while slightly reducing mIoU. In training, we use $K$=8 because of the acceptable training time and satisfactory diversity. Training with $K$=16 would take about twice as long. We discuss memory overhead and evaluation time in supplementary material.
    		
            \begin{table}[htb]
            \centering
            \def\arraystretch{1.4}
            \setlength{\tabcolsep}{0.7em} 
            \begin{tabular}{ |c|c|c|c|c| } 
            \hline
            K & mIoU & MSE & LPIPS \\
            \hline
            16 & 45.28 & 4.105 & 0.1496\\
            8 & 45.81 & 3.004 & 0.1264\\ 
            4 & 46.48 & 2.078 & 0.0947\\
            2 & 46.44 & 1.424 & 0.0608\\
            1 & 46.32 & 0.014 & 0.0010\\
            \hline
            \end{tabular}
            \vspace{1mm}
            \caption{ Impact of the number of generated samples ($K$) on mIoU and diversity, measured on mid-term prediction task and MR1 loss. We can see that a larger number of generated samples contributes to greater diversity and slightly reduces mIoU. Testing was performed at an early stage of the paper, and the results are somewhat different from those in the table \ref{tab:resmid}. Since we evaluate on only one model for each $K$, due to high variance in both training and evaluation mIoU values don't necessarily represent real situation. Furthermore, in some tasks training with $K$=12 or $K$=16 sometimes showed better accuracy. The number of generated samples listed in the table refers to the learning phase, while we measure mIoU, MSE and LPIPS on 8 generated samples in the model exploitation phase. }
            \label{tab:numsamples}
            \end{table}

    	\subsection{Importance of GAN loss} 
\begin{figure}[htb]
            \centering
            \begin{subfigure}[b]{0.9\textwidth}
              \includegraphics[width=1\linewidth]{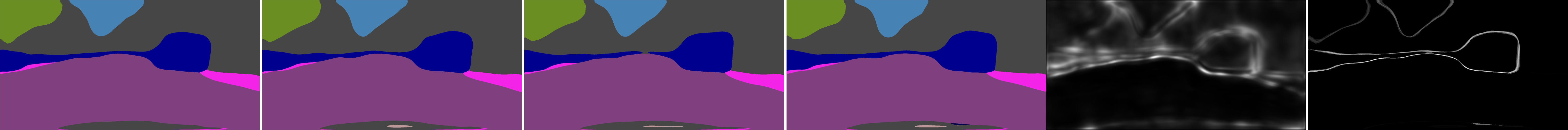}
              \includegraphics[width=1\linewidth]{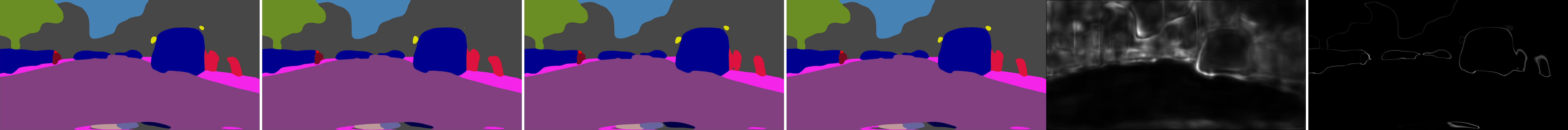}
              \includegraphics[width=1\linewidth]{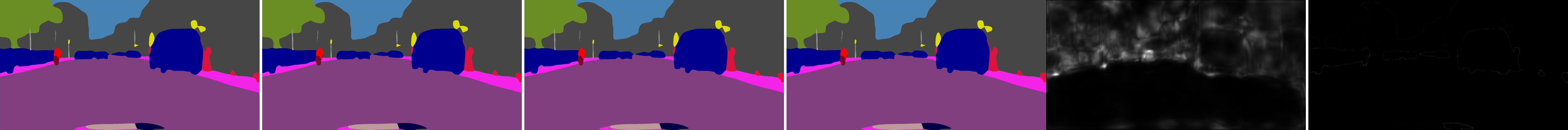}
            \caption{ $\lambda_{GAN}=0$}
            \label{fig:wogan}
            \end{subfigure}
            
            \begin{subfigure}[b]{0.9\textwidth}
              \includegraphics[width=1\linewidth]{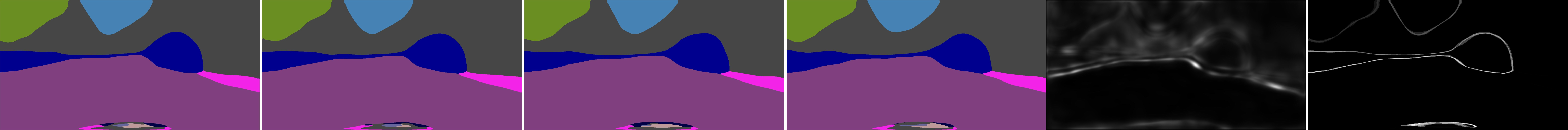}
              \includegraphics[width=1\linewidth]{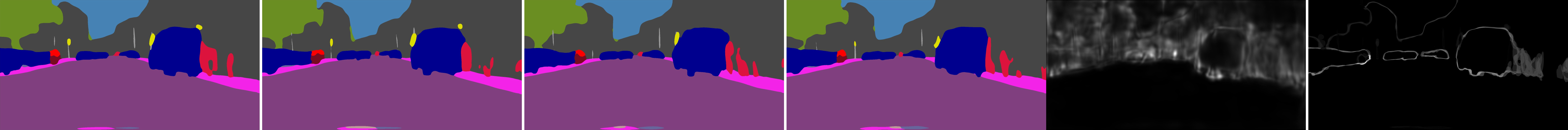}
              \includegraphics[width=1\linewidth]{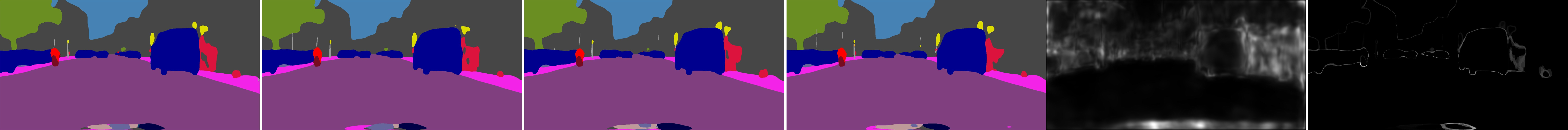}
            \caption{$\lambda_{GAN}=10$}
            \label{fig:wgan}
            \end{subfigure} 
            \caption{The first four columns 
              show 4 out of 8 generated predictions. 
              The last two columns show mean logit variance
              and the variance of discrete predictions. 
              Samples in the three rows are generated 
              in epochs 1, 13 and 161, respectively.
              On image a), in the last row we observe almost indistinguishable predictions with second gray image being entirely black.
              On image b), we see some diversity in people behind a van, and last image shows some diversity in predicted classes.
            }
            \end{figure}

            To show that the output diversity is not only due to the use of moment reconstruction losses and random noise, we trained the model without adversarial loss ($\lambda_{GAN}=0$). In this experiment we were using MR1 loss on short-term forecasting task and with generator producing 8 predictions for each input image. Although some diversity is visible at an early stage (MSE around 0.7), around the 12th epoch the diversity is less and less noticeable (MSE around 0.35), and after the 40th it can barely be seen (MSE around 0.1). The model achieved its best mIoU 59.18 in epoch 160 (although it was trained on 400 epochs), and the MSE measure in that epoch was 0.08. Figure \ref{fig:wogan} shows the improvement in performance through the epochs, but with gradual weakening of diversity. For this example we chose an image with a lot of void surfaces, due to the fact that greatest diversity is usually seen in those places. On Figure \ref{fig:wgan} we show the same scene, but predictions were obtained on a model trained with weights $\lambda_{GAN}=10$ and $\lambda_{MR}=100$. It took the model 232 epochs to achieve its best mIoU which is 59.14, but MSE held stable above 1 until the last, 400th, epoch.
            
         %\newpage
         
       	\subsection{Diversity of multiple forecasts}
		We have seen that averaging generated predictions before grading them increases mIoU - from 0.2 up to 3 percentage points, depending on the task. 
		Therefore, we propose a novel metric for measuring
		plausibility of multimodal forecasting. 
		The proposed metric measures percentage of pixels 
		that were correctly classified at least once 
		through multiple forecasts. 
		We distinguish three cases by looking at: 
		
		\begin{enumerate}
		\item Every pixel except void class
		\item Only pixels of movable objects
		\item Only pixels that were correctly classified by Oracle.
		\end{enumerate}  
		
\begin{figure}[htb]
	        \centering
            \includegraphics[width=1\linewidth]{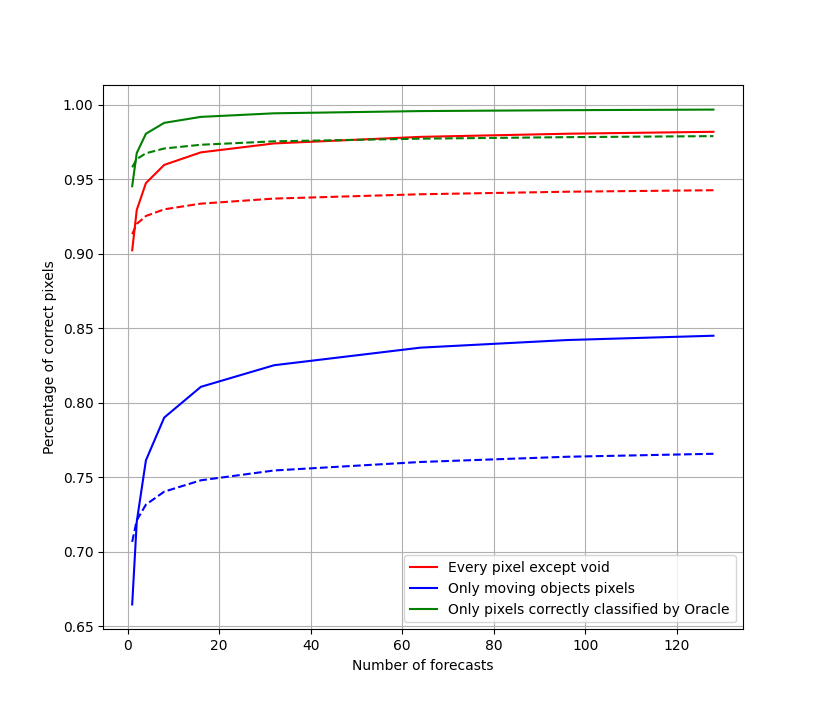}
            \caption{Number of future pixels
            that were correctly classified at least once depending on the number of forecasts. We show three different cases with lines of different colors, as described by legend. Full lines represent MR2 short-term model, while dashed lines represent results with MR1 short-term model.
            }
            \label{fig:divmetr}
        \end{figure}

		We measure at multiple checkpoints (1, 2, 4, ..., 128) and present the obtained results in Figure \ref{fig:divmetr}. On Figure \ref{fig:divmetr} we compare short term forecasting using MR1 and MR2 losses, while in supplementary material we also show additional line which represents \textit{best so far} prediction.
		
    %\clearpage	
    \section{Conclusion and Future Work}
    
    We have presented a novel approach
    for multimodal semantic forecasting 
    in road driving scenarios. 
    Our approach achieves multi-modality
    by injecting random noise 
    into the feature forecasting module.
    Hence, the feature forecast module becomes
    a conditional feature-to-feature generator
    which is trained by minimizing 
    the moment reconstruction loss,
    and by maximizing the loss of a patch-level discriminator.
    Both the generator and the discriminator 
    operate on abstract features.
    
    We have also proposed a novel metric
    for measuring plausibility 
    of multimodal forecasting.
    The proposed metric measures
    the number of forecasts
    required to correctly guess
    a given proportion of all future pixels.
    We encourage the metric to 
    reflect forecasting performance
    by disregarding pixels 
    which are not correctly guessed by the oracle.
    
    The inference speed of our multi-modal model 
    is similar to the uni-modal baseline. 
    Experiments show that the proposed setup 
    is able to achieve 
    considerable diversity in mid-term forecasting.
    MR2 loss brings more diversity compared to MR1, 
    however it reduces mIoU by around 5 percentage points.
    Inspection of the generated forecasts,
    reveals that the model is sometimes still 
    hesitating to replace close and large objects,
    but it often accepts to take a chance 
    on close and dynamic or small and distant objects, 
    like bikes and pedestrians.
    
    In the future work we shall consider training
    with \textit{proxy MR1} and \textit{proxy MR2} losses. 
    We should also consider using different discriminator,
    for example the one with global contextual information.
    Also, one of the options is to try concatenating features with their 
    spatial pools prior to the discriminator.
    Other suitable future directions include 
    evaluating performance on the instance segmentation task 
    and experimenting with different generative models. 

	\bibliographystyle{splncs04}
	\bibliography{egbib}

\begin{thebibliography}{10}
\providecommand{\url}[1]{\texttt{#1}}
\providecommand{\urlprefix}{URL }
\providecommand{\doi}[1]{https://doi.org/#1}

\bibitem{bhattacharyya2018bayesian}
Bhattacharyya, A., Fritz, M., Schiele, B.: Bayesian prediction of future street
  scenes using synthetic likelihoods. In: International Conference on Learning
  Representations (2019)

\bibitem{chen20arxiv}
Chen, L., Lopes, R.G., Cheng, B., Collins, M.D., Cubuk, E.D., Zoph, B., Adam,
  H., Shlens, J.: Leveraging semi-supervised learning in video sequences for
  urban scene segmentation. CoRR  \textbf{abs/2005.10266} (2020)

\bibitem{gan}
Goodfellow, I.J., Pouget-Abadie, J., Mirza, M., Xu, B., Warde-Farley, D.,
  Ozair, S., Courville, A., Bengio, Y.: Generative adversarial networks (2014)

\bibitem{probabilistic}
Hu, A., Cotter, F., Mohan, N., Gurau, C., Kendall, A.: Probabilistic future
  prediction for video scene understanding (2020)

\bibitem{adlearning}
Hung, W., Tsai, Y., Liou, Y., Lin, Y., Yang, M.: Adversarial learning for
  semi-supervised semantic segmentation. In: British Machine Vision Conference
  2018, {BMVC} 2018, Northumbria University, Newcastle, UK, September 3-6,
  2018. p.~65 (2018)

\bibitem{pix2pix}
Isola, P., Zhu, J.Y., Zhou, T., Efros, A.A.: Image-to-image translation with
  conditional adversarial networks (2016)

\bibitem{kirillov2018panoptic}
Kirillov, A., He, K., Girshick, R., Rother, C., Doll{\'a}r, P.: Panoptic
  segmentation. arXiv preprint arXiv:1801.00868  (2018)

\bibitem{prostate}
Kohl, S., Bonekamp, D., Schlemmer, H.P., Yaqubi, K., Hohenfellner, M.,
  Hadaschik, B., Radtke, J.P., Maier-Hein, K.: Adversarial networks for the
  detection of aggressive prostate cancer (2017)

\bibitem{kosaraju19nips}
Kosaraju, V., Sadeghian, A., Mart\'{\i}n-Mart\'{\i}n, R., Reid, I.,
  Rezatofighi, H., Savarese, S.: Social-bigat: Multimodal trajectory
  forecasting using bicycle-gan and graph attention networks. In: Advances in
  Neural Information Processing Systems 32, pp. 137--146 (2019)

\bibitem{ladder}
Krešo, I., Krapac, J., Šegvić, S.: Efficient ladder-style {DenseNets} for
  semantic segmentation of large images. IEEE Transactions on Intelligent
  Transportatoin Systems  (2020)

\bibitem{supres}
Ledig, C., Theis, L., Huszar, F., Caballero, J., Cunningham, A., Acosta, A.,
  Aitken, A., Tejani, A., Totz, J., Wang, Z., Shi, W.: Photo-realistic single
  image super-resolution using a generative adversarial network (2016)

\bibitem{mrgan}
Lee, S., Ha, J., Kim, G.: Harmonizing maximum likelihood with gans for
  multimodal conditional generation (2019)

\bibitem{patchgan}
Li, C., Wand, M.: Precomputed real-time texture synthesis with markovian
  generative adversarial networks (2016)

\bibitem{vidpred}
Liang, X., Lee, L., Dai, W., Xing, E.P.: Dual motion gan for future-flow
  embedded video prediction. In: Proceedings of the IEEE International
  Conference on Computer Vision. pp. 1744--1752 (2017)

\bibitem{f2f}
Luc, P., Couprie, C., LeCun, Y., Verbeek, J.: Predicting future instance
  segmentation by forecasting convolutional features (2018)

\bibitem{luc2017predicting}
Luc, P., Neverova, N., Couprie, C., Verbeek, J., LeCun, Y.: Predicting deeper
  into the future of semantic segmentation. In: Proceedings of the IEEE
  International Conference on Computer Vision. pp. 648--657 (2017)

\bibitem{rprior}
Makansi, O., Cicek, O., Buchicchio, K., Brox, T.: Multimodal future
  localization and emergence prediction for objects in egocentric view with a
  reachability prior. In: The IEEE/CVF Conference on Computer Vision and
  Pattern Recognition (CVPR) (June 2020)

\bibitem{cgan}
Mirza, M., Osindero, S.: Conditional generative adversarial nets (2014)

\bibitem{saric}
{\v{S}}ari{\'c}, J., Or{\v{s}}i{\'c}, M., Antunovi{\'c}, T., Vra{\v{z}}i{\'c},
  S., {\v{S}}egvi{\'c}, S.: Single level feature-to-feature forecasting with
  deformable convolutions. In: German Conference on Pattern Recognition. pp.
  189--202. Springer (2019)

\bibitem{saric20cvpr}
Saric, J., Orsic, M., Antunovic, T., Vrazic, S., Segvic, S.: Warp to the
  future: Joint forecasting of features and feature motion. In: 2020 {IEEE/CVF}
  Conference on Computer Vision and Pattern Recognition, {CVPR} 2020, Seattle,
  WA, USA, June 13-19, 2020. pp. 10645--10654. {IEEE} (2020)

\bibitem{su2017predicting}
Su, S., Pyo~Hong, J., Shi, J., Soo~Park, H.: Predicting behaviors of basketball
  players from first person videos. In: Proceedings of the IEEE Conference on
  Computer Vision and Pattern Recognition. pp. 1501--1510 (2017)

\bibitem{sun19mm}
Sun, J., Xie, J., Hu, J., Lin, Z., Lai, J., Zeng, W., Zheng, W.: Predicting
  future instance segmentation with contextual pyramid convlstms. In: ACM MM.
  pp. 2043--2051 (2019)

\bibitem{terwilliger2019recurrent}
Terwilliger, A., Brazil, G., Liu, X.: Recurrent flow-guided semantic
  forecasting. In: 2019 IEEE Winter Conference on Applications of Computer
  Vision (WACV). pp. 1703--1712. IEEE (2019)

\bibitem{vondrick2015anticipating}
Vondrick, C., Pirsiavash, H., Torralba, A.: Anticipating the future by watching
  unlabeled video. arXiv preprint arXiv:1504.08023  \textbf{2} (2015)

\bibitem{vukotic2017one}
Vukoti{\'c}, V., Pintea, S.L., Raymond, C., Gravier, G., van Gemert, J.C.:
  One-step time-dependent future video frame prediction with a convolutional
  encoder-decoder neural network. In: International Conference on Image
  Analysis and Processing. pp. 140--151. Springer (2017)

\bibitem{yang18cvpr}
Yang, M., Yu, K., Zhang, C., Li, Z., Yang, K.: Denseaspp for semantic
  segmentation in street scenes. In: CVPR. pp. 3684--3692 (2018)

\bibitem{yao19icra}
Yao, Y., Xu, M., Choi, C., Crandall, D.J., Atkins, E.M., Dariush, B.:
  Egocentric vision-based future vehicle localization for intelligent driving
  assistance systems. In: ICRA (2019)

\bibitem{inpaintning}
Yu, J., Lin, Z., Yang, J., Shen, X., Lu, X., Huang, T.S.: Generative image
  inpainting with contextual attention. In: The IEEE Conference on Computer
  Vision and Pattern Recognition (CVPR) (June 2018)

\bibitem{lpips}
Zhang, R., Isola, P., Efros, A.A., Shechtman, E., Wang, O.: The unreasonable
  effectiveness of deep features as a perceptual metric (2018)

\bibitem{zhao17cvpr}
Zhao, H., Shi, J., Qi, X., Wang, X., Jia, J.: Pyramid scene parsing network.
  In: ICCV (2017)

\bibitem{zhen19aaai}
Zhen, M., Wang, J., Zhou, L., Fang, T., Quan, L.: Learning fully dense neural
  networks for image semantic segmentation. In: {AAAI} (2019)

\end{thebibliography}
	
\end{document}

\pagestyle{headings}
	\mainmatter

	% Insert your submission number here
	\def\GCPR20SubNumber{104}

	% Replace with your title
	\title{Multimodal semantic forecasting based on conditional generation of future features - Supplementary Material}

	% DO NOT MODIFY these for the draft version that is used for the
	% review process.
	\titlerunning{Multimodal semantic forecasting based on conditional generation of future...}
	\authorrunning{Kristijan Fugošić, Josip Šarić and Siniša Šegvić}
	\author{Kristijan Fugošić, Josip Šarić and Siniša Šegvić}
	\institute{University of Zagreb Faculty of Electrical Engineering and Computing, Croatia}

	\maketitle
	\section{Experimental results}
	       \begin{figure}[H]
            \centering
            \includegraphics[width=1\linewidth]{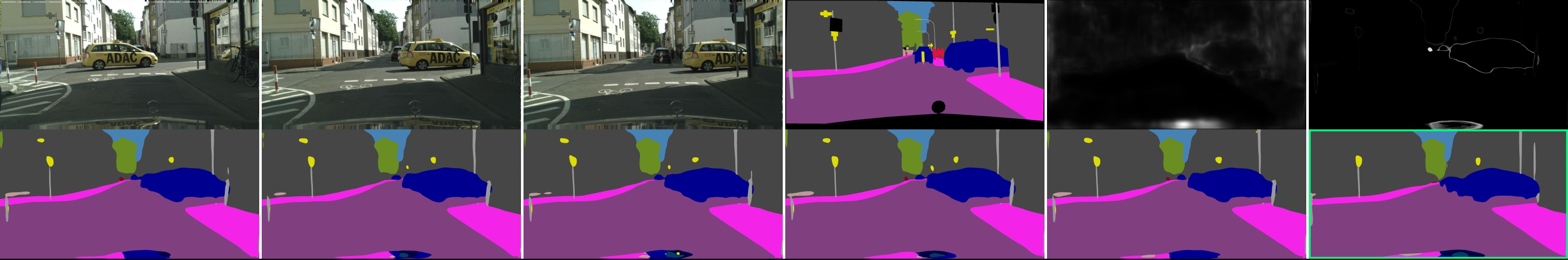}
            \includegraphics[width=1\linewidth]{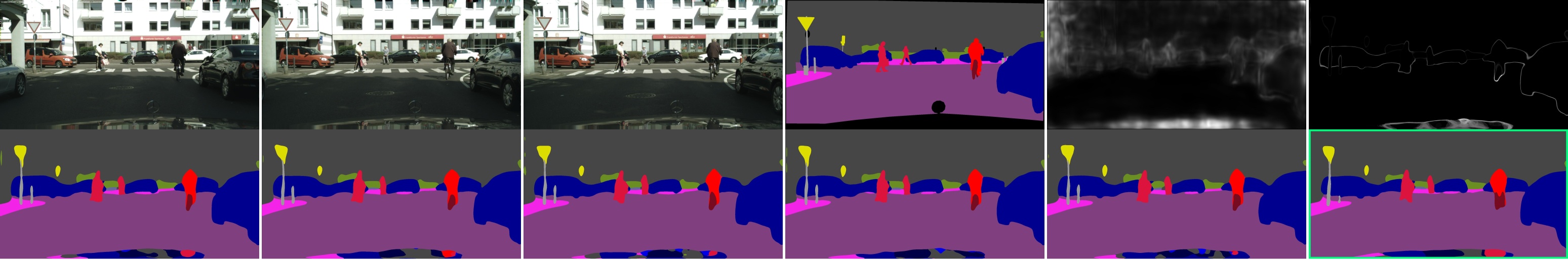}
            \includegraphics[width=1\linewidth]{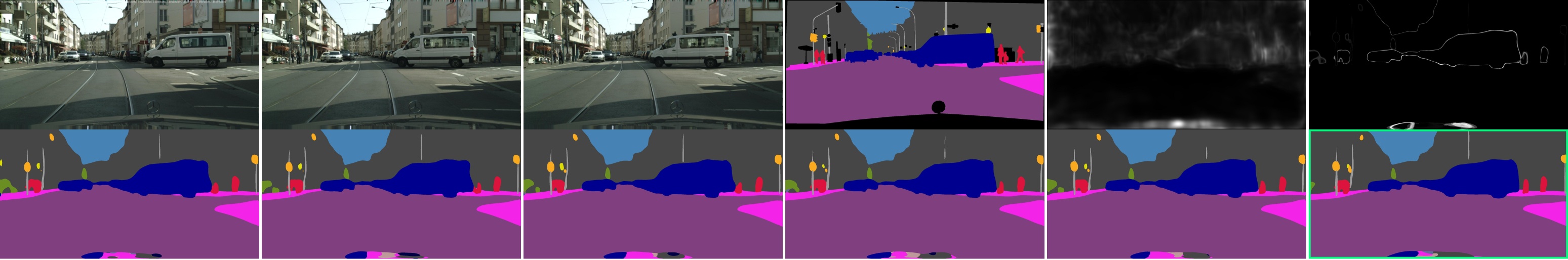}
            \includegraphics[width=1\linewidth]{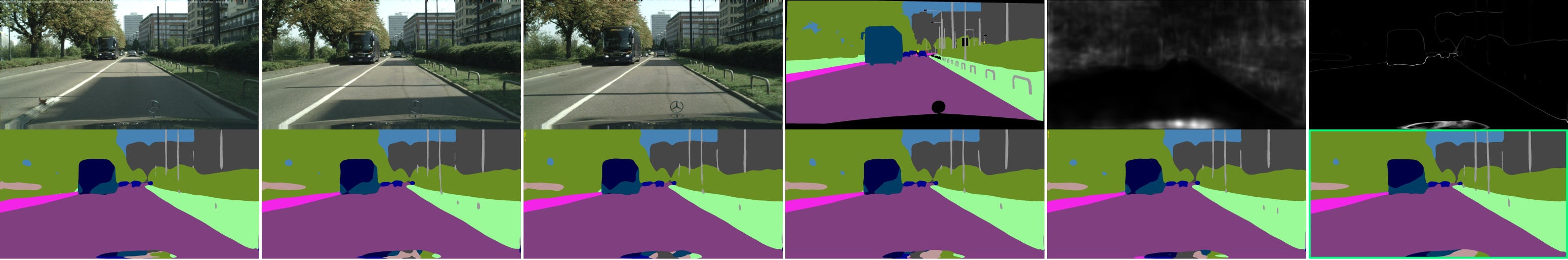}
            \includegraphics[width=1\linewidth]{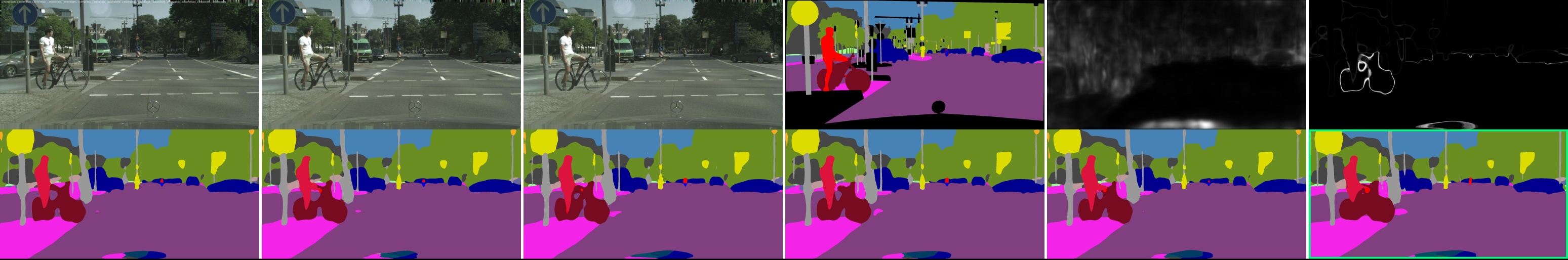}
            \includegraphics[width=1\linewidth]{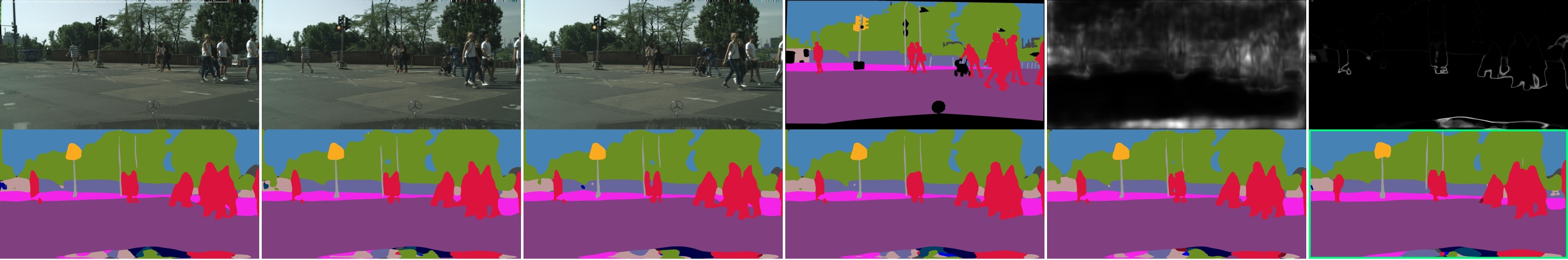}
            \includegraphics[width=1\linewidth]{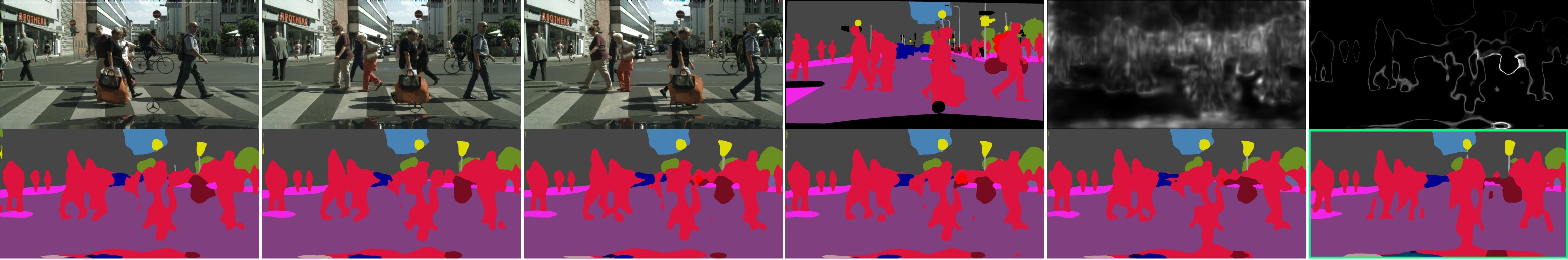}
            \includegraphics[width=1\linewidth]{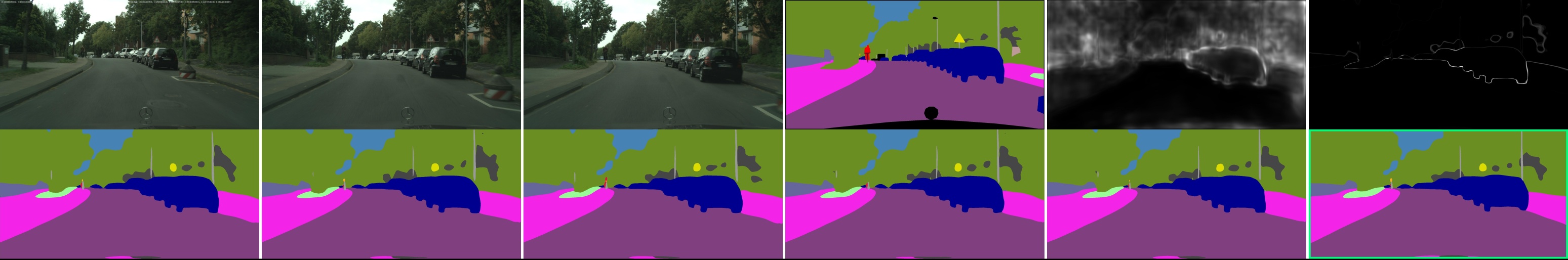}
            \label{fig:smr1}
            \caption{\textbf{Short-term forecasting using MR1 loss.} For every scene, in the first row we show: first and last frame that model has observed, future frame and its ground-truth segmentation, mean logit variance and variance of discrete predictions. In second row we show 5 of our predictions and baseline model prediction (mIoU 58.98) at the end (green frame).}
            \end{figure}

            \newpage
            \begin{figure}[H]
            \centering
            \includegraphics[width=1\linewidth]{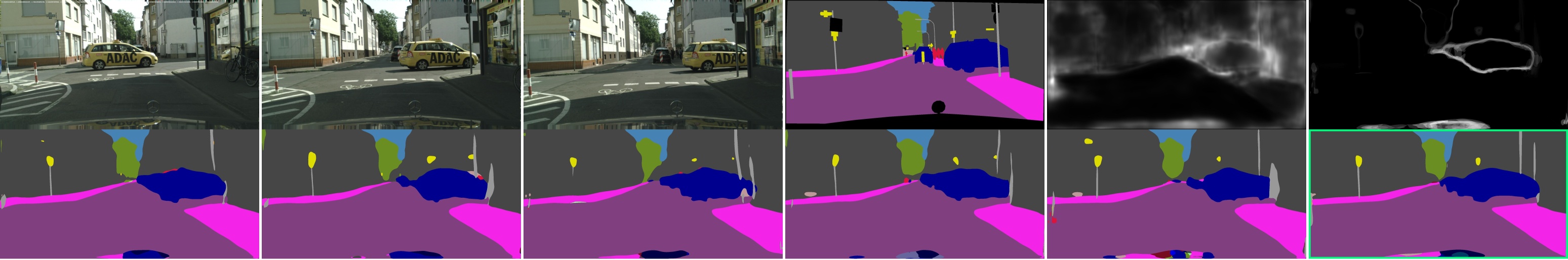}
            \includegraphics[width=1\linewidth]{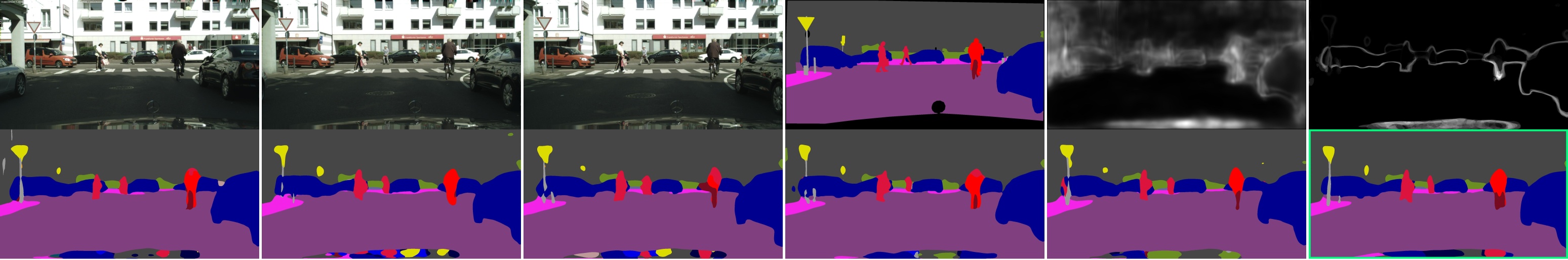}
            \includegraphics[width=1\linewidth]{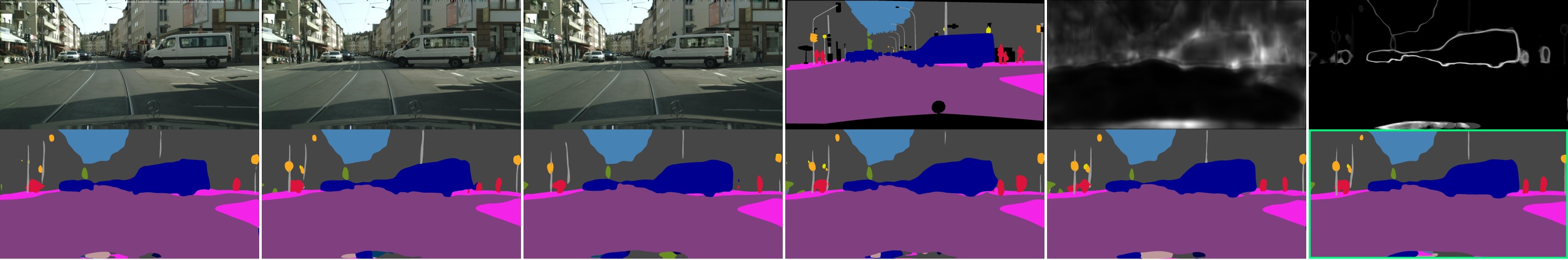}
            \includegraphics[width=1\linewidth]{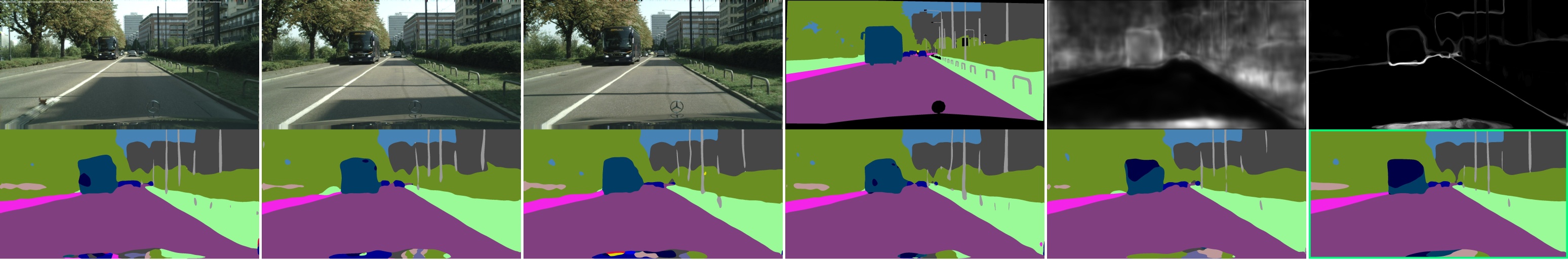}
            \includegraphics[width=1\linewidth]{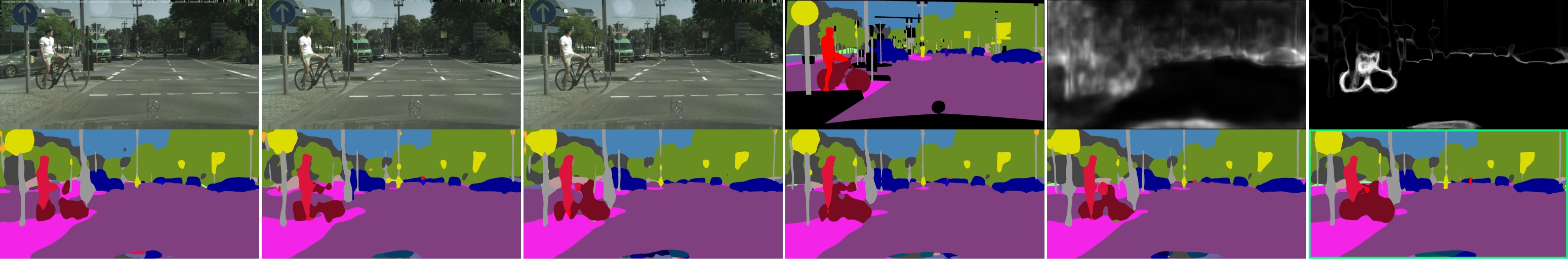}
            \includegraphics[width=1\linewidth]{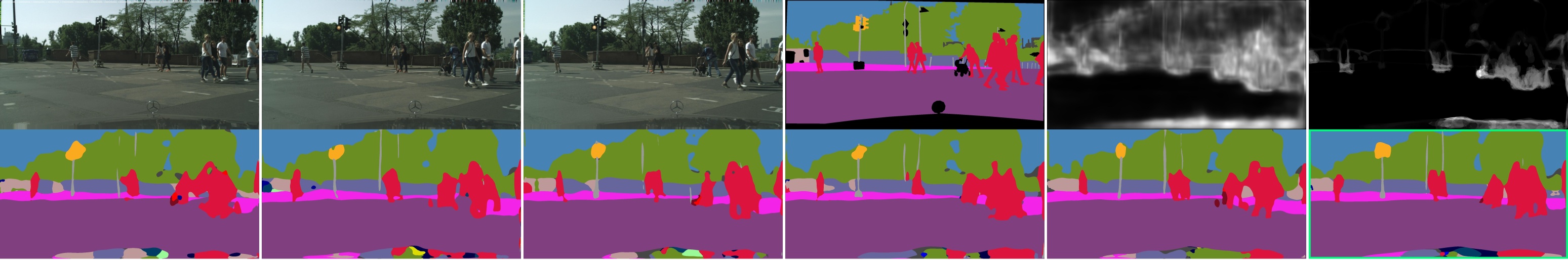}
            \includegraphics[width=1\linewidth]{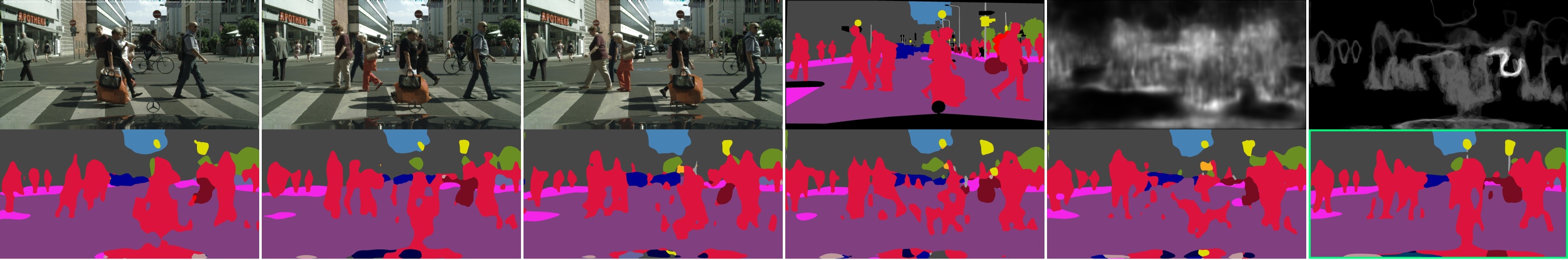}
            \includegraphics[width=1\linewidth]{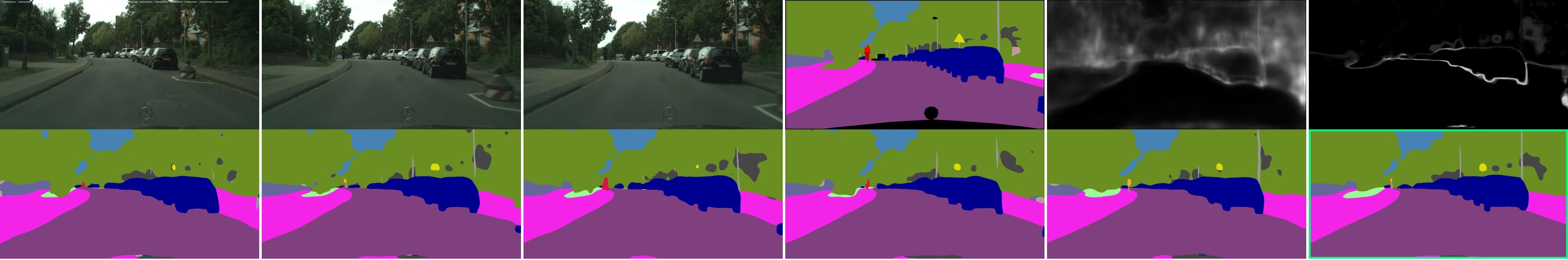}
            \label{fig:smr2}
            \caption{\textbf{Short-term forecasting using MR2 loss.} For every scene, in the first row we show: first and last frame that model has observed, future frame and its ground-truth segmentation, mean logit variance and variance of discrete predictions. In second row we show 5 of our predictions and baseline model prediction (mIoU 58.98) at the end (green frame).}
            \end{figure}
            
            \newpage
            \begin{figure}[H]
            \centering
            \includegraphics[width=1\linewidth]{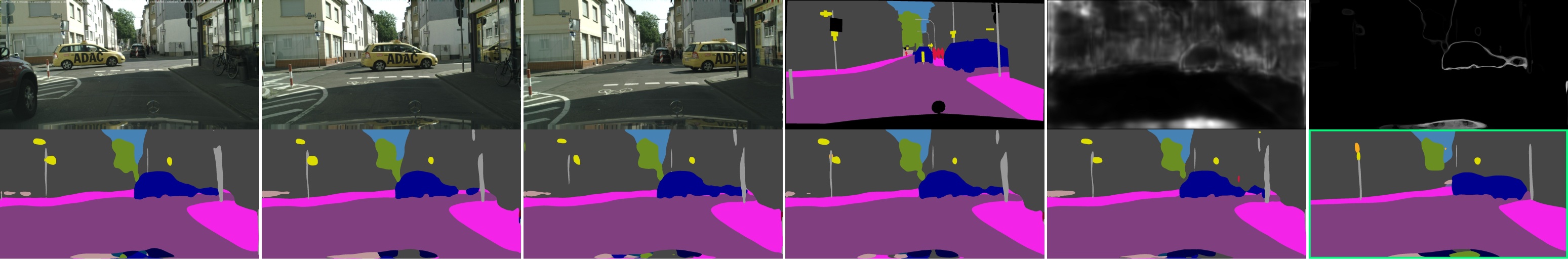}
            \includegraphics[width=1\linewidth]{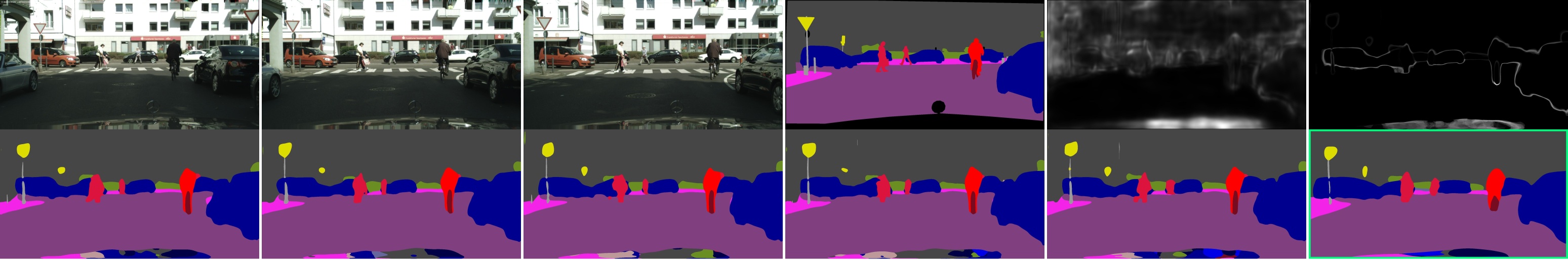}
            \includegraphics[width=1\linewidth]{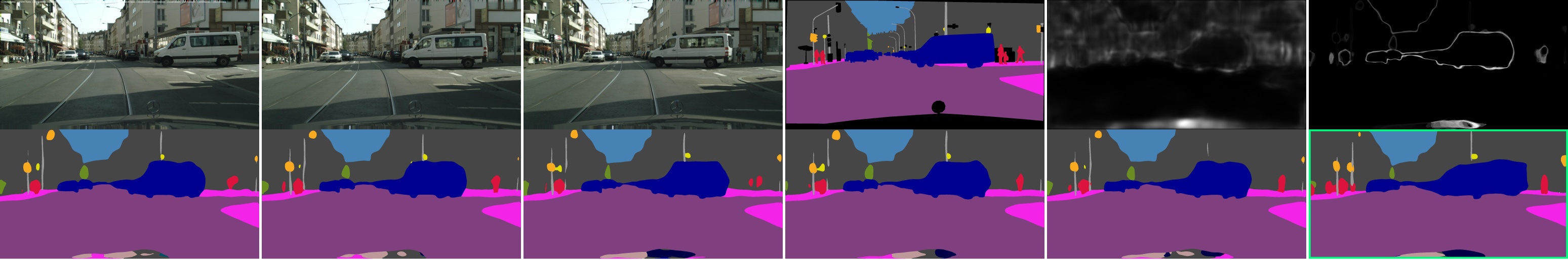}
            \includegraphics[width=1\linewidth]{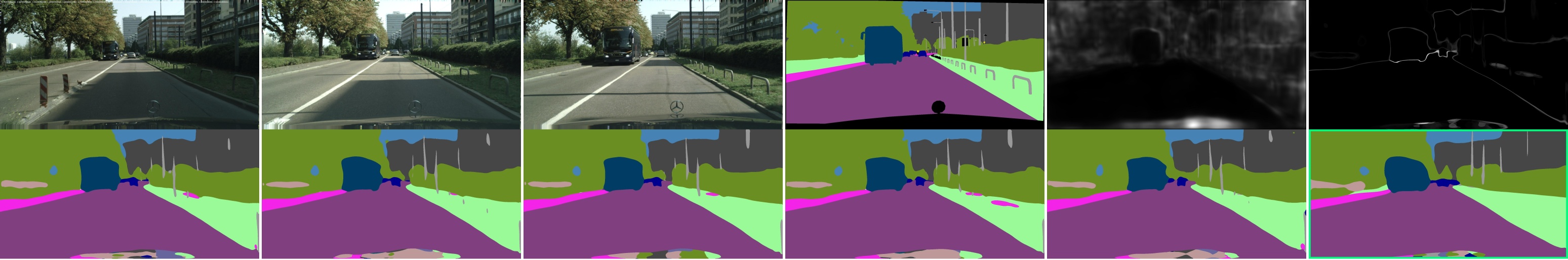}
            \includegraphics[width=1\linewidth]{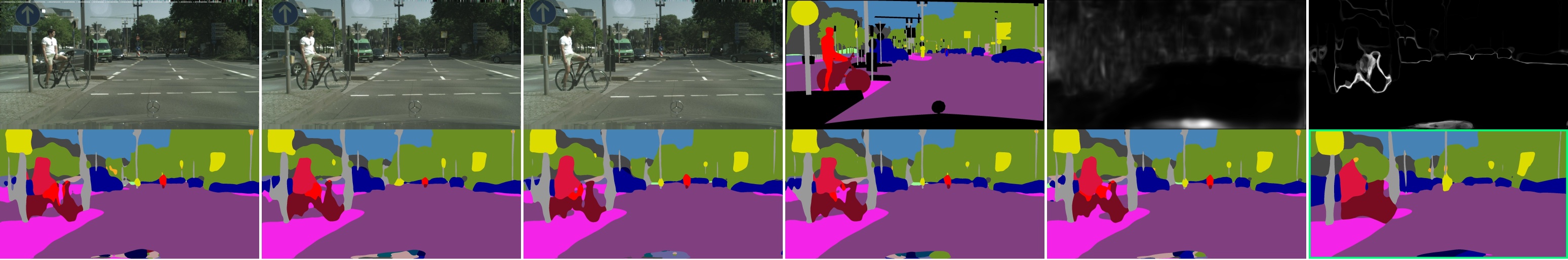}
            \includegraphics[width=1\linewidth]{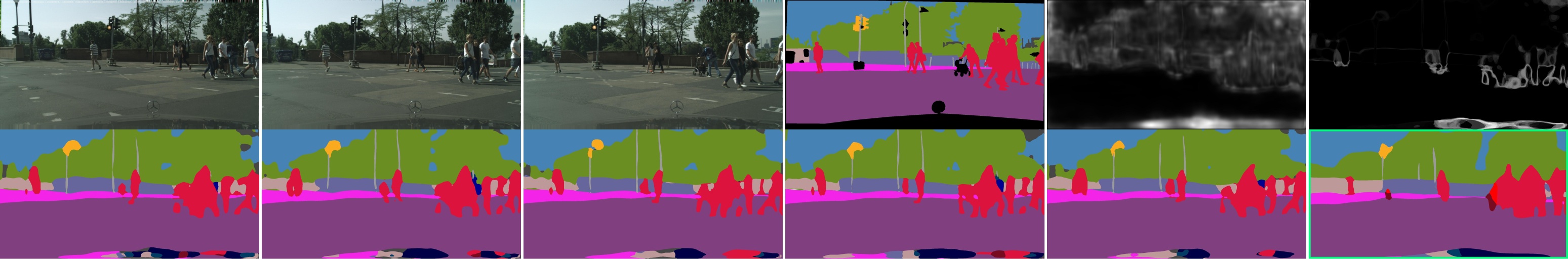}
            \includegraphics[width=1\linewidth]{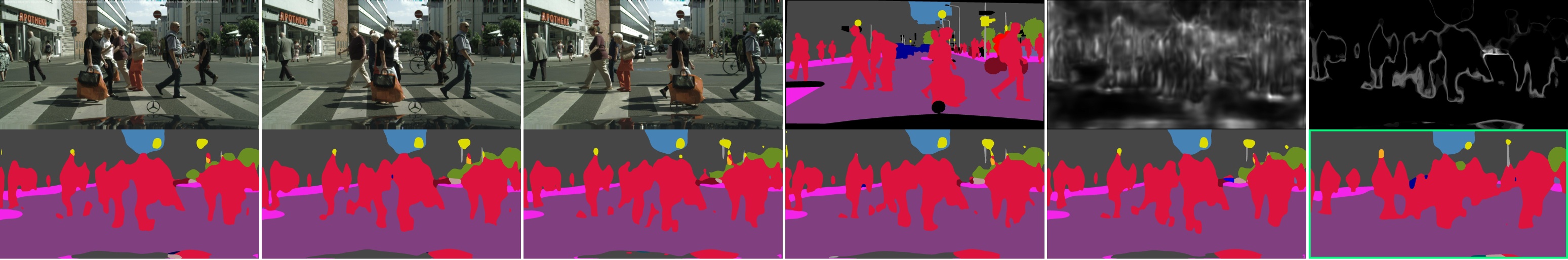}
            \includegraphics[width=1\linewidth]{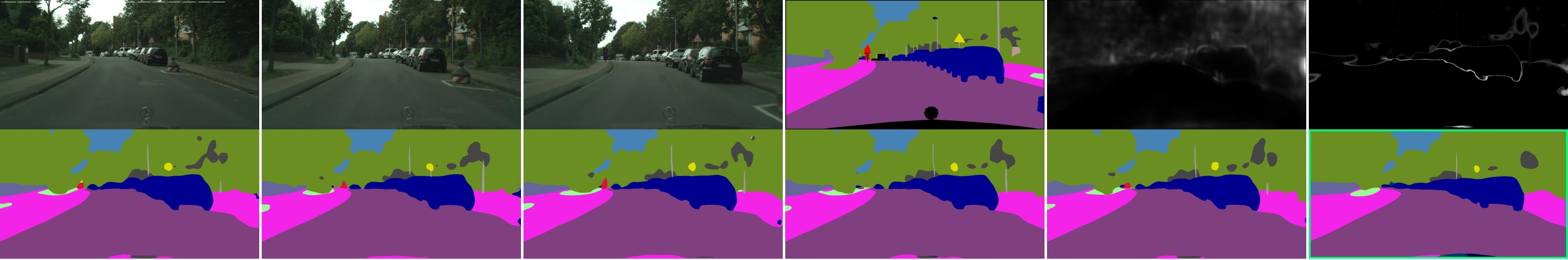}
            \label{fig:mmr1}
            \caption{\textbf{Mid-term forecasting using MR1 loss.} For every scene, in the first row we show: first and last frame that model has observed, future frame and its ground-truth segmentation, mean logit variance and variance of discrete predictions. In second row we show 5 of our predictions and baseline model prediction (mIoU 46.68) at the end (green frame).}
            \end{figure}
            
            \newpage
            \begin{figure}[H]
            \centering
            \includegraphics[width=1\linewidth]{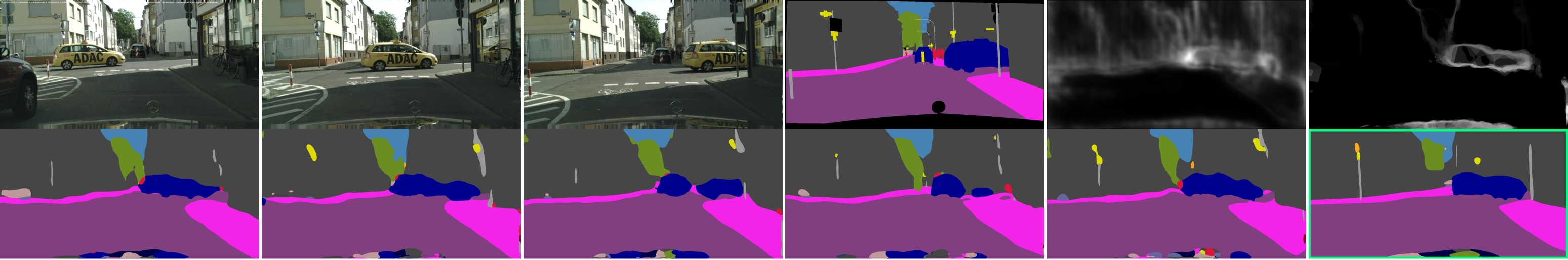}
            \includegraphics[width=1\linewidth]{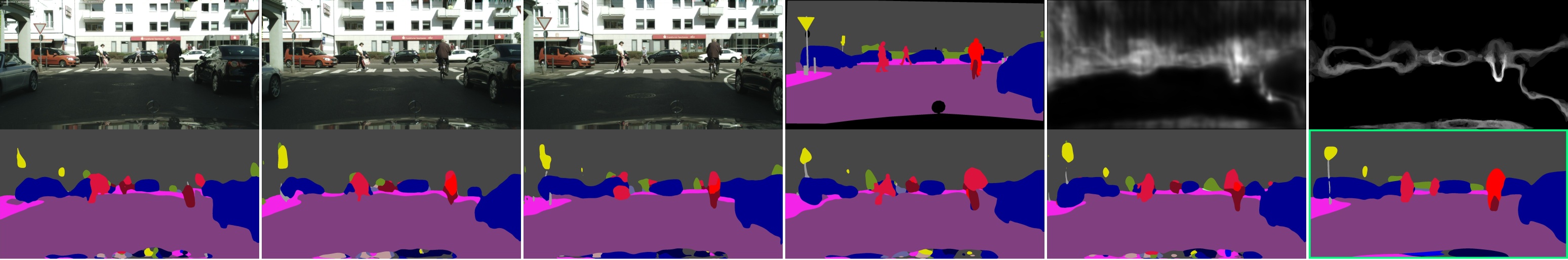}
            \includegraphics[width=1\linewidth]{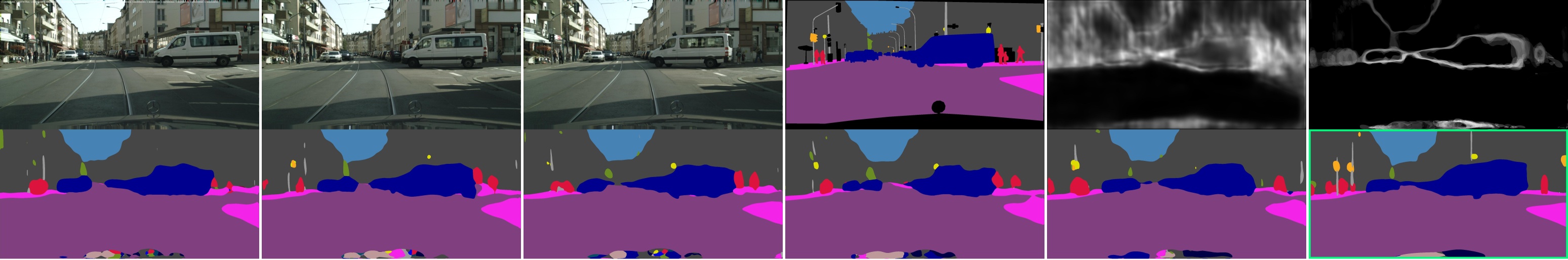}
            \includegraphics[width=1\linewidth]{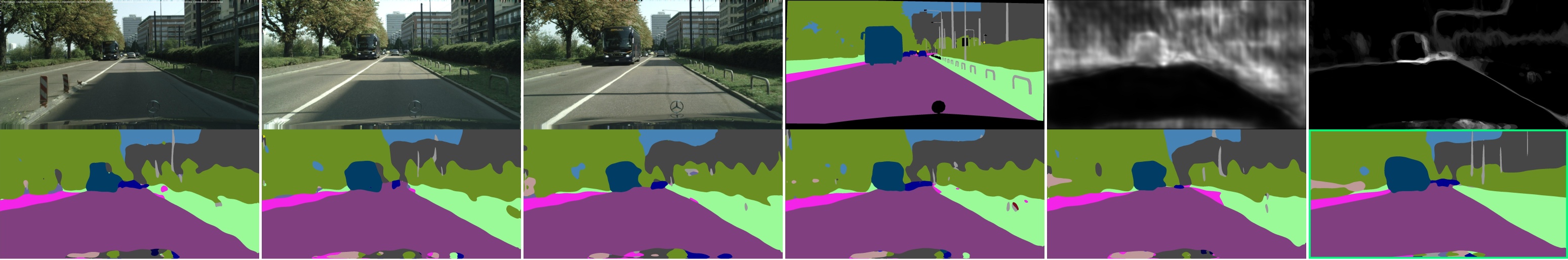}
            \includegraphics[width=1\linewidth]{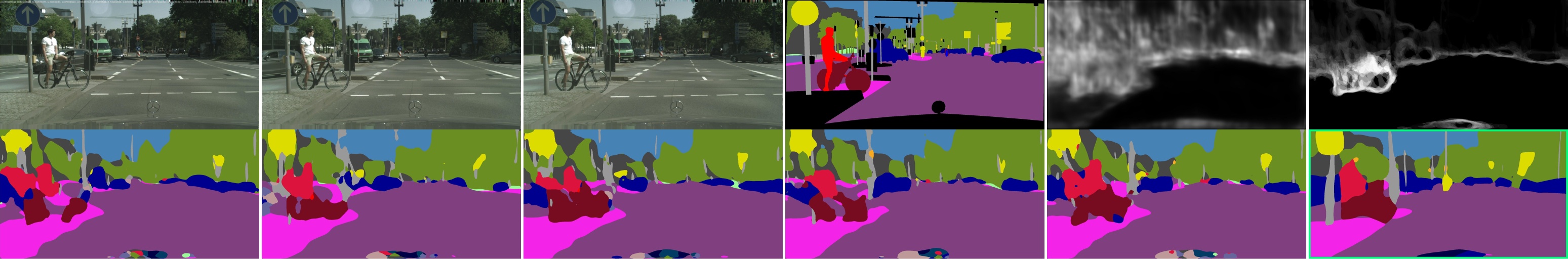}
            \includegraphics[width=1\linewidth]{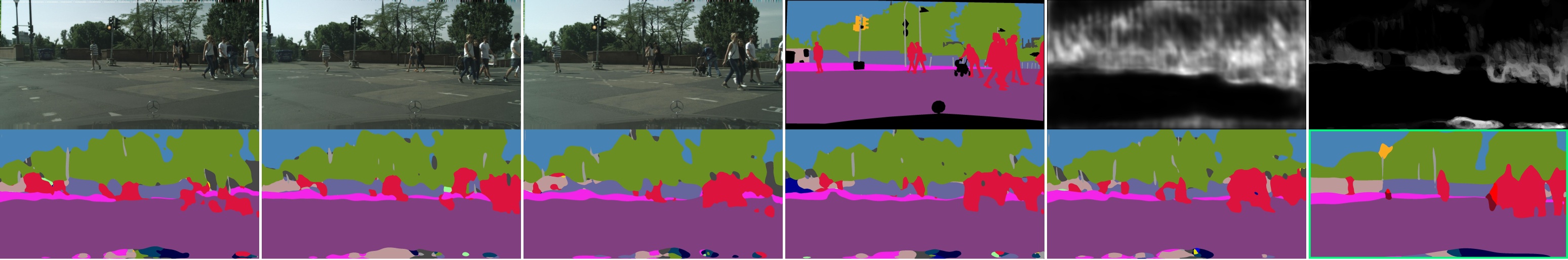}
            \includegraphics[width=1\linewidth]{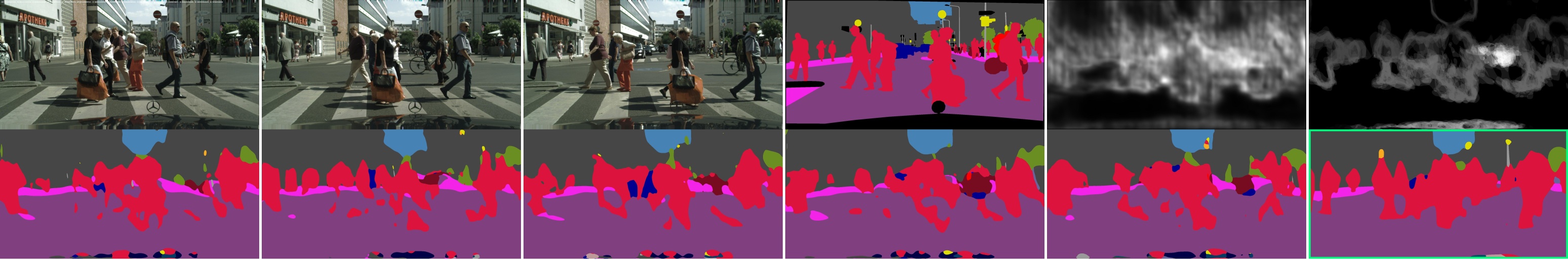}
            \includegraphics[width=1\linewidth]{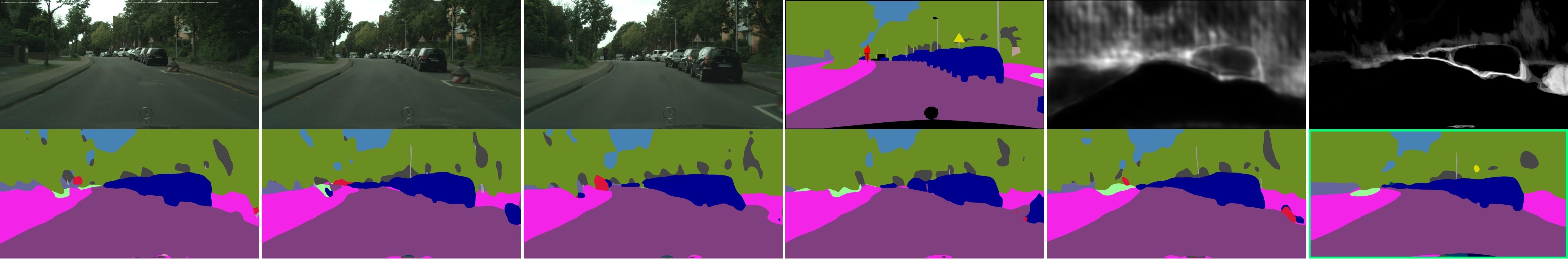}
            \label{fig:mmr2}
            \caption{\textbf{Mid-term forecasting using MR2 loss.} For every scene, in the first row we show: first and last frame that model has observed, future frame and its ground-truth segmentation, mean logit variance and variance of discrete predictions. In second row we show 5 of our predictions and baseline model prediction (mIoU 46.68) at the end (green frame).}
            \end{figure}

            \newpage
    
            \subsection{Individual, averaged and top 5\% predictions comparison}

    In Table \ref{tab:additionalm} we compare top 5\% of predictions with predictions that were averaged before evaluation. Measurement was done with K=20, which means that top 5\% is the best image out of 20. We can see once again that using MR1 loss results in predictions of higher quality, even better than the averaged predictions. It should be noted that measurements were taken on only one trained model for each task, so when it comes to model performance, numbers in the main paper are more credible.
    
            \begin{table} [htb] 
                \centering
                \def\arraystretch{1.2}
                \setlength{\tabcolsep}{0.7em}           
                \begin{tabular}{ |c|c|c|c|c|c| } 
                \hline
                Method & Type & Time & Loss & mIoU & mIoU-MO \\
                \hline
                MM-DeformF2F-8 & Normal & Short-term & MR1 & 59.27$^{\pm 0.006}$ & 56.27$^{\pm 0.0.013}$ \\
                MM-DeformF2F-8 & Average & Short-term & MR1  & 59.57$^{\pm 0.007}$ &56.59$^{\pm 0.015}$ \\
                MM-DeformF2F-8 & Top 5\% & Short-term & MR1 & \textbf{60.10}$^{\pm 0.042}$ & \textbf{57.26}$^{\pm 0.100}$ \\
                \hline
                MM-DeformF2F-8 & Normal & Short-term & MR2 & 54.35$^{\pm 0.052}$ & 50.34$^{\pm 0.110}$ \\
                MM-DeformF2F-8 & Average & Short-term & MR2 & \textbf{57.47}$^{\pm 0.044}$ & \textbf{54.48}$^{\pm 0.109}$ \\
                MM-DeformF2F-8 & Top 5\% & Short-term & MR2 & 57.41$^{\pm 0.202}$ & 54.44$^{\pm 0.422}$ \\
                \hline
                MM-DeformF2F-8 & Normal & Mid-term & MR1 & 46.48$^{\pm 0.004}$ & 41.45$^{\pm 0.012}$ \\
                MM-DeformF2F-8 & Average & Mid-term & MR1  & 46.92$^{\pm 0.010}$ & 41.86$^{\pm 0.017}$ \\
                MM-DeformF2F-8 & Top 5\% & Mid-term & MR1 & \textbf{47.51}$^{\pm 0.072}$ & \textbf{42.56}$^{\pm 0.155}$ \\
                \hline
                MM-DeformF2F-8 & Normal & Mid-term & MR2 & 37.80$^{\pm 0.0.013}$ & 29.28$^{\pm 0.043}$ \\
                MM-DeformF2F-8 & Average & Mid-term & MR2 & \textbf{40.64}$^{\pm 0.033}$ & \textbf{32.44}$^{\pm 0.085}$ \\
                MM-DeformF2F-8 & Top 5\% & Mid-term & MR2 & 40.36$^{\pm 0.06}$ & 32.24$^{\pm 0.140}$ \\
                \hline
                \end{tabular}
                \vspace{1mm}
                \caption{Short and mid-term comparison of top 5\% and averaged predictions, K=20}
                \label{tab:additionalm}
            \end{table}

        \subsection{Memory overhead and evaluation time}
        With the increase in the number of generated predictions (K), we also observe increase in memory and time required for evaluation. However, as we can see in Table \ref{tab:memtime}, increment isn't K-fold. Passing trough feature extractor takes roughly the same amount of time, while F2F model observes slight increase. Noticeable difference is in the upsampling branch. But, when we evaluate dataset of 500 images on GTX980, baseline model usually takes around two minutes and ten seconds, while evaluation on our model with K=8 takes around 15 seconds longer, which isn't as large overhead as one might expect. 
        
            \begin{table}[htb]
                \centering
                \def\arraystretch{1.2}
                \setlength{\tabcolsep}{0.7em} 
                \begin{tabular}{ |c|c|c|c|c|c| } 
                \hline
                K & F2F pass (ms) & Upsampling (ms) & Max memory allocated (MB) \\
                \hline
                1 & 8 & 13 & 531\\
                8 & 17 & 75 & 3342\\
                16 & 31 & 184 & 6610\\
                \hline
                \end{tabular}
                \vspace{1mm}
                \caption{Time and memory required to pass trough F2F and upsampling path (GTX1070)  for different K.}
                \label{tab:memtime}
            \end{table}
        
        Total time includes half-sized images being passed trough feature extractor (although during the training we load extracted features from SSD).
        
        \newpage
        
       	\subsection{Diversity of multiple forecasts}
\begin{figure}[htb] 
              \begin{subfigure}[b]{1\linewidth}
                \centering
                \includegraphics[width=0.8\linewidth]{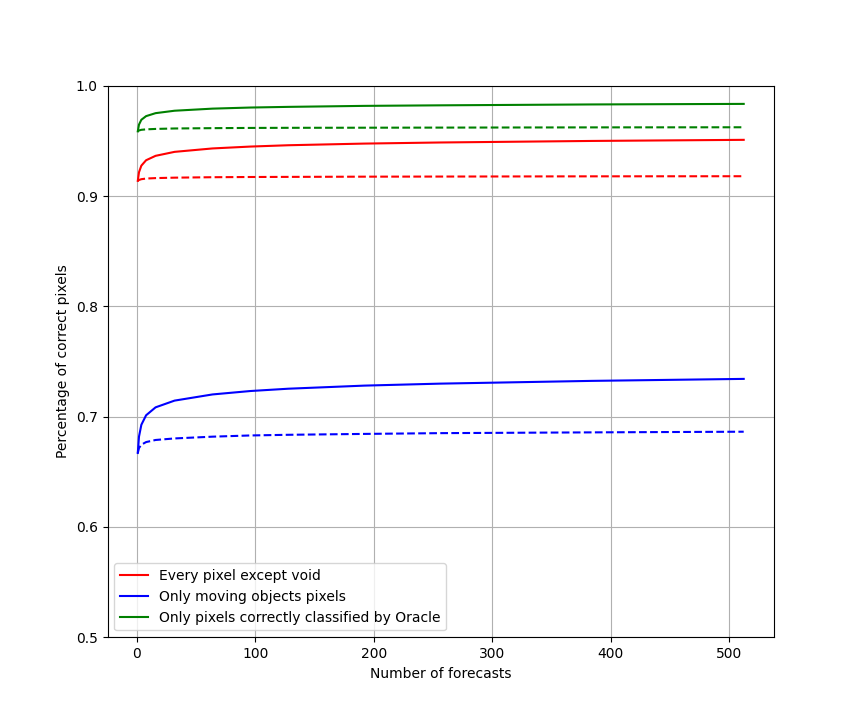} 
                \caption{Short-term MR1} 
              \end{subfigure}
              \begin{subfigure}[b]{1\linewidth}
                \centering
                \includegraphics[width=0.8\linewidth]{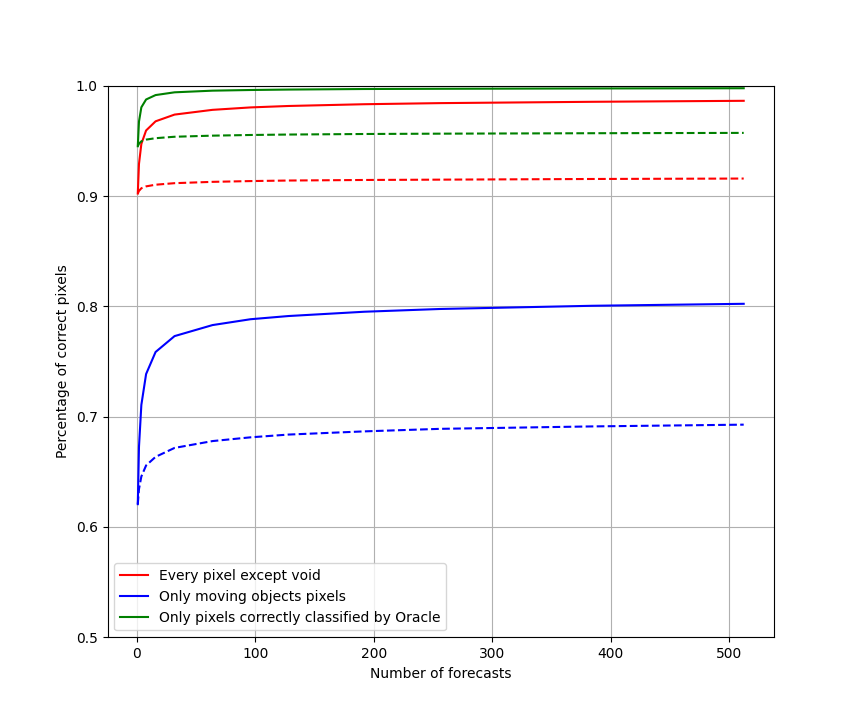} 
                \caption{Short-term MR2} 
              \end{subfigure} 
              \caption{\textbf{Short-term.} With full line we show percentage of pixels that were correctly classified at least once depending on the number of forecasts. With dashed line we show single best percentage of pixels that were correctly classified as of yet. We show three different cases with lines of different colors, as described by legend. }
              \label{diversitim1} 
            \end{figure}

\begin{figure}[htb] 
              \begin{subfigure}[b]{1\linewidth}
                \centering
                \includegraphics[width=0.8\linewidth]{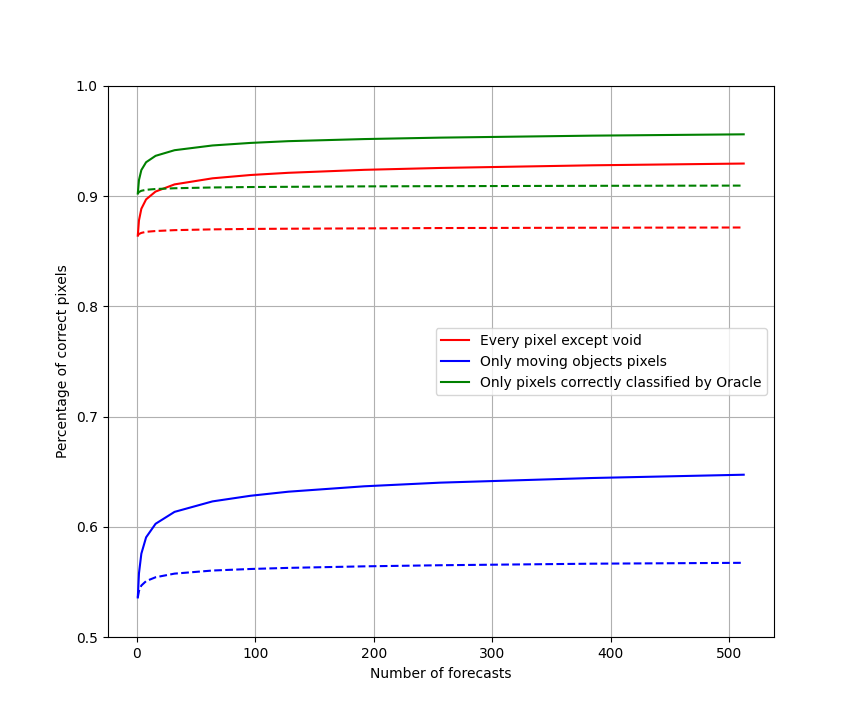} 
                \caption{Mid-term MR1} 
              \end{subfigure}
              \begin{subfigure}[b]{1\linewidth}
                \centering
                \includegraphics[width=0.8\linewidth]{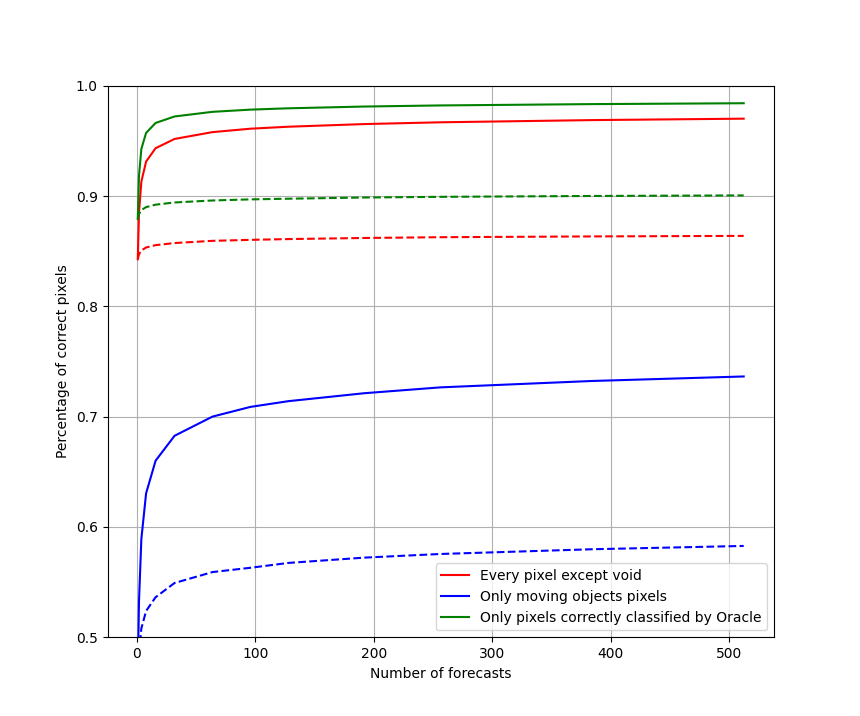} 
                \caption{Mid-term MR2} 
              \end{subfigure} 
              \caption{\textbf{Mid-term.} With full line we show percentage of pixels that were correctly classified at least once depending on the number of forecasts. With dashed line we show single best percentage of pixels that were correctly classified as of yet. We show three different cases with lines of different colors, as described by legend. }
              \label{diversitim2} 
            \end{figure}

        On figures \ref{diversitim1} and \ref{diversitim2} we can see percentage of pixels that were correctly classified at least once trough multiple forecasts. We also show single best percentage as of yet with dashed lines. If we compare MR1 and MR2, we can see that MR2 rises faster and peaks higher, which is a sign of greater diversity. We can also observe faster growth if we compare mid-term graphs with their short-term counterparts, like we expected.
        
